\documentclass{article}

\usepackage{iclr2027_conference,times}

\usepackage{microtype}
\usepackage{graphicx}
\usepackage{subcaption}
\usepackage{booktabs}
\PassOptionsToPackage{hyphens}{url}
\usepackage{hyperref}
\usepackage{url}
\usepackage{algorithm}
\usepackage{algorithmic}
\usepackage{amsmath}
\usepackage{amssymb}
\usepackage{mathtools}
\usepackage{amsthm}
\usepackage{xcolor}
\usepackage[capitalize,noabbrev]{cleveref}
\usepackage{multirow}
\usepackage{wrapfig}
\usepackage{tabularx} 
\usepackage{needspace}

\theoremstyle{plain}

\theoremstyle{definition}

\theoremstyle{remark}

\title{Improving Video Sparse Attention\\with Fine-grained Router and Sparse Rebasing}

\author{
\begin{tabular}{@{}l@{}}
\textbf{Peiyuan Zhang\textsuperscript{1,$\dagger$},
Guoqiang Wei\textsuperscript{2},
Yilong Zhao\textsuperscript{3,$\dagger$},
Zixiang Zhang\textsuperscript{2},
Wei Zhou\textsuperscript{4},}\\
\textbf{Will Lin\textsuperscript{1},
Heng Zhang\textsuperscript{2},
Xiaonan Nie\textsuperscript{2},
Yan Zeng\textsuperscript{2},
Hao Zhang\textsuperscript{1}}
\end{tabular}\\[0.5em]
\textsuperscript{1}University of California, San Diego\\
\textsuperscript{2}ByteDance Seed\\
\textsuperscript{3}University of California, Berkeley\\
\textsuperscript{4}Georgia Institute of Technology\\
\textsuperscript{$\dagger$} Work done during an internship at ByteDance Seed
}

\iclrfinalcopy 

\hypersetup{hidelinks,pdfkeywords={Machine Learning, ICML}}

\newcommand{\routerillustration}{%
  \begingroup
  \setlength{\unitlength}{\dimexpr\linewidth/2884\relax}%
  \begin{picture}(2884,1908)
    \put(0,0){\includegraphics[width=\linewidth]{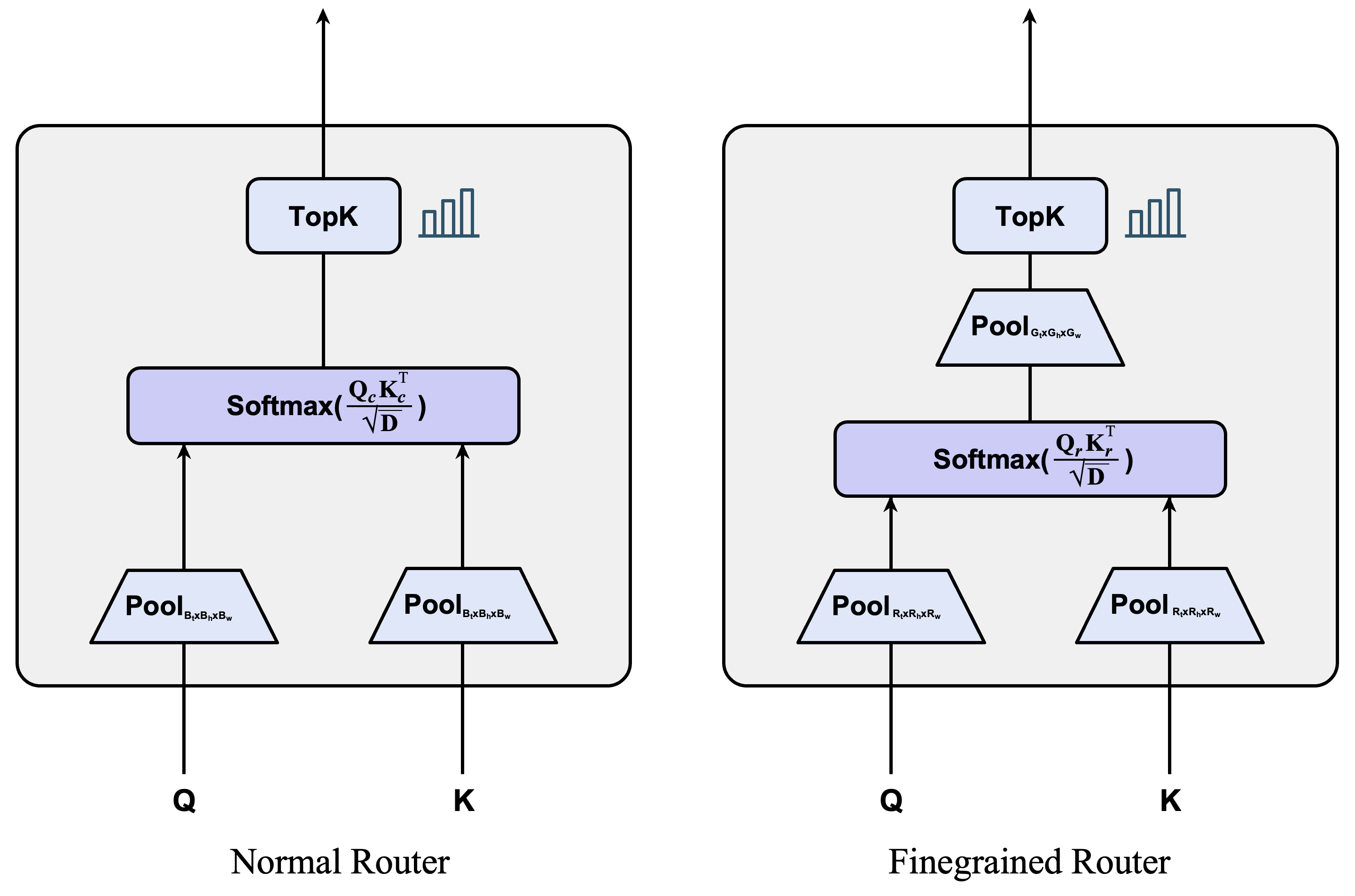}}
    \put(1790,20){\color{white}\rule{1060\unitlength}{110\unitlength}}
    \put(1790,20){\makebox(1060,110){\fontsize{5.1}{6}\selectfont Fine-grained Router}}
  \end{picture}%
  \endgroup
}

\newcommand{\ablationillustration}{%
  \begingroup
  \setlength{\unitlength}{\dimexpr\linewidth/7246\relax}%
  \begin{picture}(7246,3446)
    \put(0,0){\includegraphics[width=\linewidth]{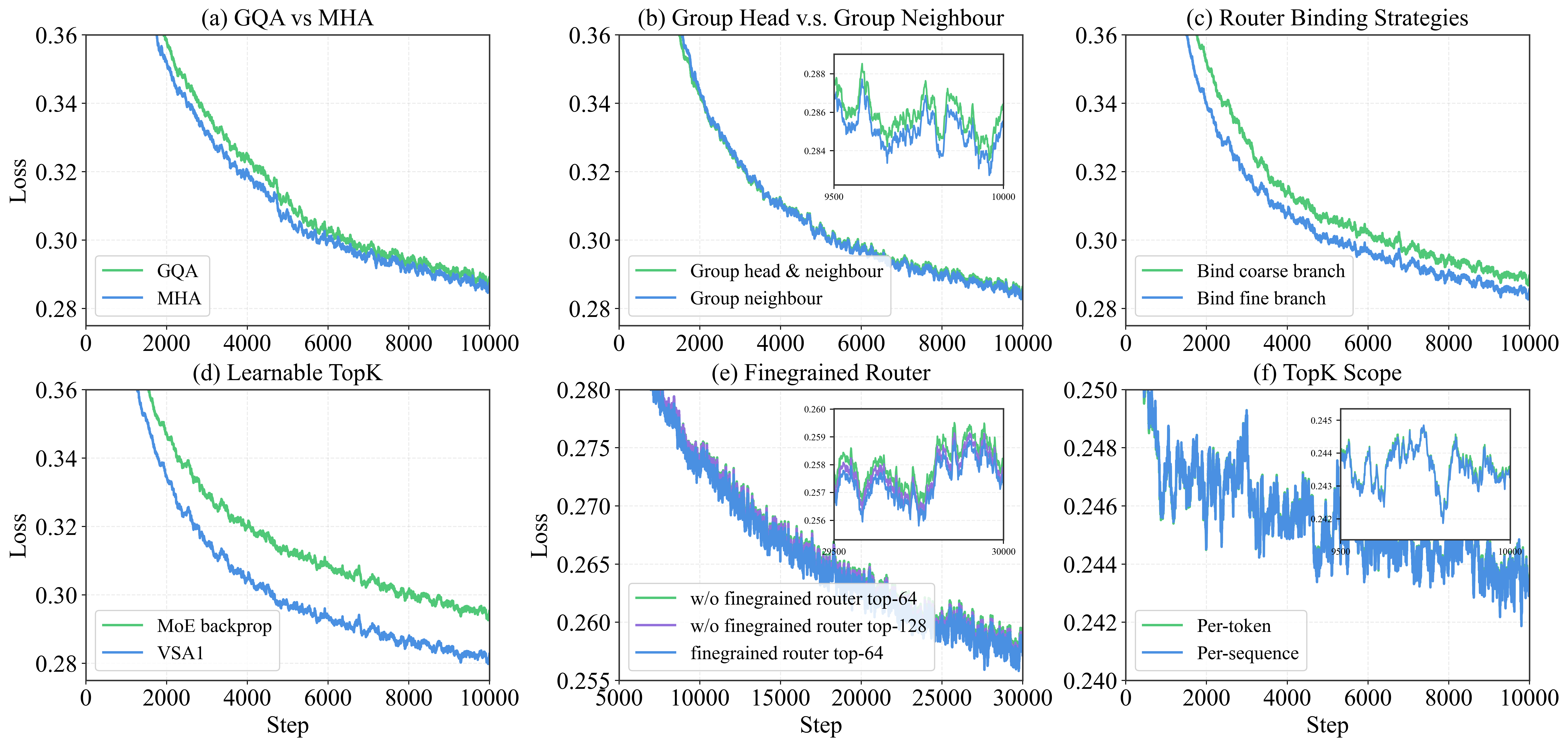}}
    \put(750,3298){\color{white}\rule{1200\unitlength}{148\unitlength}}
    \put(750,3298){\makebox(1200,148){\fontsize{6.3}{7}\selectfont (a) GQA vs. MHA}}
    \put(2900,3298){\color{white}\rule{1770\unitlength}{148\unitlength}}
    \put(2900,3298){\makebox(1770,148){\fontsize{6.3}{7}\selectfont (b) Group Head vs. Group Neighbour}}
    \put(3200,1655){\color{white}\rule{1250\unitlength}{140\unitlength}}
    \put(3200,1655){\makebox(1250,140){\fontsize{6.3}{7}\selectfont (e) Fine-grained Router}}
    \put(3175,621){\color{white}\rule{1125\unitlength}{116\unitlength}}
    \put(3175,621){\makebox(1125,116)[l]{\fontsize{5}{6}\selectfont w/o fine-grained router top-64}}
    \put(3175,493){\color{white}\rule{1125\unitlength}{116\unitlength}}
    \put(3175,493){\makebox(1125,116)[l]{\fontsize{5}{6}\selectfont w/o fine-grained router top-128}}
    \put(3175,367){\color{white}\rule{1125\unitlength}{116\unitlength}}
    \put(3175,367){\makebox(1125,116)[l]{\fontsize{5}{6}\selectfont fine-grained router top-64}}
  \end{picture}%
  \endgroup
}

\begin{document}

\newcommand{\methodname}{Video Sparse Attention 2}
\newcommand{\methodnameshort}{\textsc{VSA2}}

\newcommand{\curriculumname}{\textsc{Hard-to-Easy Curriculum}}
\newcommand{\upcyclingname}{\textsc{Sparse Rebasing}}
\newcommand{\red}[1]{{\color{red}#1}}
\newcommand{\TODO}[1]{\textbf{\color{red}[TODO: #1]}}

\providecommand{\maketitlesupplementary}{}

\maketitle
\lhead{Preprint}

\begin{abstract}

We present \methodnameshort, a frontier trainable sparse attention for video DiTs. \methodnameshort~includes a variety of new architectural features and training procedures that we apply across all stages of the DiT development cycle -- including pretraining, RL, and inference -- to produce a DiT with comparable or better quality than a full attention counterpart. Architecturally, \methodnameshort~introduces a fine-grained router that improves the precision of identifying critical tokens and supports dynamic computation by allowing each query to attend to a variable number of key–value pairs. 
In training, we identify a Hard-to-Easy Curriculum, where models trained under high sparsity and later evaluated with lower sparsity during inference not only generalize effectively, but also outperform models trained with full attention in motion quality. \methodnameshort~is also flexible: it can replace full attention during the middle of progressive low-to-high resolution pretraining, rebasing early-stage full-attention checkpoints. Experiments show that \methodnameshort~reduces attention computation by half over VSA with lower loss. On 720p videos, it accelerates attention by 8.9$\times$ and end-to-end generation by 4.62$\times$ compared to the FlashAttention-3 baseline, while achieving comparable or better video quality.
\end{abstract}
\vspace{-1em}
\section{Introduction}
\label{sec:intro}

Video diffusion transformers (DiTs) are rapidly approaching practical utility. Recent systems, such as Genie 3~\citep{genie3}, generate interactive, minute-long, high-resolution video streams while maintaining persistent memory of earlier events. Scaling video DiTs to this regime is nontrivial: even a 5-second HD clip exceeds 100K tokens; this makes self-attention, whose cost is quadratic with sequence length, the dominating cost at both training and inference~\citep{yang2024cogvideox,kong2024hunyuanvideo, wan2025wan}. Practically, most entries in $\mathrm{Softmax}(\mathbf{QK}^\top/\sqrt{\mathbf{D}})$ contribute negligibly to attention output, with a small set of \emph{critical tokens} that carry the signal~\citep{zhang2023h2o, jiang2024minference, ding2025efficient}. This observation has driven a wave of \emph{inference-only} sparse attention methods~\citep{zhang2025sparge,xi2025sparsevideo, xu2025xattention,zhang2025fast} that prune low-weight interactions to accelerate DiT. While effective for inference, these methods leave the costly pretraining phase untouched and thus cannot unlock long-context training.

To reduce \emph{training} FLOPs, the field has turned to trainable sparse attention~\cite{lu2025moba,zhang2025faster,zhan2025bidirectional,zhang2025sla}. 
A common design is a two-stage, coarse-to-fine framework: a coarse branch acts as a router to pool neighboring tokens into tiles and computes full attention at tile granularity to form a coarse attention map; based on the map, the coarse branch applies TopK to identify tiles that likely contain critical tokens. Then, a fine branch computes token-level block-sparse attention only within the selected tiles. 
These methods have managed to sparsify full attention by $80\%$ at post-training, but run into a sparsity ceiling for two mechanism-level reasons.


First, for hardware efficiency, the fine branch must use block-sparse kernels with block size $B$ (e.g., 64, 128) aligned with hardware characteristics. In most prior routers, the pooling stride of the coarse branch implicitly matches $B$, so each coarse token represents a $(B_t{\times}B_h{\times}B_w)$ cube. This coarse view blurs the sparse structures in $\mathbf{QK}^\top$. At long horizon and high resolution, this aliasing either hurts the recall of truly critical tokens, results in a quality drop, or forces more tiles to be kept, leading to a sparsity drop.
Second, standard per-token topK forces every query to attend to the same number of KVs, regardless of difficulty. Easy queries waste compute on redundant context; hard queries are under-provisioned. This caps performance at high sparsity because raising sparsity uniformly starves the very queries that need more context.


We present \methodname~(\methodnameshort), a sparse attention mechanism that allows for greater sparsity and more precise critical token identification. This is enabled by a finer-grained router decoupled from hardware block size, and a sequence-level budget that is fixed overall but adaptively allocated across queries. As shown in Figure~\ref{fig:router_comparison} and Table~\ref{tab:vsa2_notation}, \methodnameshort~decouples the mean-pooling size of the router from hardware-efficient block size. We use a hardware-efficient block size $B$ for sparse attention
and a smaller pooling size $R$ for the router's queries and keys.
After softmax, we aggregate the router scores using
$G \times G$ pooling, where $G=B/R$.
In parallel, we replace per-token topK with per-sequence topK, keeping the sequence-level computation the same but allowing us to assign a variable compute budget for different query tokens.
With a fine-grained router and per-sequence topK, \methodnameshort~achieves lower loss than full attention while being 90\% to 95\% sparse.


Our systematic studies on pretraining with \methodnameshort~further identify an effective training recipe named \upcyclingname. We find that \methodnameshort~functions as a drop-in replacement for full attention within a progressive low-to-high resolution training pipeline: reuse full-attention checkpoints from the early image and low-resolution video training stages, and only introduce \methodnameshort~in the most compute-demanding phases involving high-resolution, long-duration videos, avoiding training from scratch. A crucial component of \upcyclingname~is \curriculumname, where we find that training models with aggressive sparsity and later relaxing sparsity at inference leads to better motion quality compared to the full-attention baseline. We train a video DiT on 480--720p videos and RL preference pairs to verify the recipe end-to-end. The final model matches its full attention counterpart in human evaluations and sometimes outperforms it by producing better motion. At 720p, \methodnameshort~accelerates the attention operation by 8.9$\times$ and the end-to-end generation by 4.62$\times$. Further, despite being trained on 5--12\,s clips, the model directly generates 30\,s videos, indicating headroom toward minute-long generation.

In summary, this paper makes the following contributions: (1) We propose \methodnameshort, an improved video sparse attention with a fine-grained router. (2) We identify practical training recipes, \upcyclingname, for pretraining with \methodnameshort. (3) Powered by \methodnameshort, we train a video DiT that matches or outperforms the full attention counterpart while being up to 95\% sparse. To our knowledge, \methodnameshort~is the first trainable sparse attention that is end-to-end verified in various stages of video DiT development.

\begin{figure}[t]
\centering

\begin{minipage}[t]{0.48\columnwidth}
    \vspace{0pt} 
    \centering
    \routerillustration
    \captionof{figure}{
        Illustration of the sparse attention router used in
        \cite{zhang2025faster,zhang2025sparge,
        zhan2025bidirectional,zhang2025sla} versus the
        fine-grained router in our \methodnameshort.
    }
    \label{fig:router_comparison}
\end{minipage}
\hfill
\begin{minipage}[t]{0.48\columnwidth}
    \vspace{0pt}
    \centering
    \scriptsize
    \setlength{\tabcolsep}{3pt}
    \renewcommand{\arraystretch}{1.12}

    \begin{tabularx}{\linewidth}{
        @{}l>{\raggedright\arraybackslash}X@{}
    }
        \toprule
        \textbf{Symbol} & \textbf{Description} \\
        \midrule
        $T,H,W$ &
        Video dimensions after VAE compression. \\
        $L$ &
        Total number of video tokens.
        \newline $L=T\times H\times W$. \\
        $B=B_tB_hB_w$ &
        Block size and corresponding 3D cube size. \\
        $N=N_tN_hN_w$ &
        Number of cubes along each dimension.
        \newline
        $N_t=T/B_t$, $N_h=H/B_h$,
        $N_w=W/B_w$. \\
        $R=R_tR_hR_w$ &
        Router input-pooling size.
        \newline
        $R_t<B_t$, $R_h<B_h$, $R_w<B_w$. \\
        $G=G_tG_hG_w$ &
        Router score-pooling size.
        \newline
        $G_t=B_t/R_t$, $G_h=B_h/R_h$,
        $G_w=B_w/R_w$. \\
        \bottomrule
    \end{tabularx}

    \captionof{table}{Notation summary for \methodname.}
    \label{tab:vsa2_notation}
\end{minipage}

\end{figure}

\section{Method}


\subsection{Background}
\label{sec:background}

Most existing video DiTs employ 3D full attention to capture dependencies across the entire spatiotemporal volume. A video latent of shape $(T, H, W)$ is first flattened into a 1D sequence of length $L = THW$, and then the attention output is computed as: $\mathbf{S} = \mathbf{Q} \mathbf{K}^\top/\sqrt{D}, 
\mathbf{P} = \text{Softmax}(\mathbf{S} + \mathbf{M}), 
\mathbf{O} = \mathbf{P} \mathbf{V}.
$
In \emph{full attention}, all entries in $\mathbf{M}$ are zero, allowing dense interactions throughout the sequence. \emph{Sparse attention} instead introduces $-\infty$ entries in $\mathbf{M}$, which can reduce FLOPs by skipping the corresponding computation in both $\mathbf{Q}\mathbf{K}^\top$ and $\mathbf{P}\mathbf{V}$. 

Since attention scores are highly non-uniform, full attention can be approximated by sparse attention if we construct a mask $\mathbf{M}$ that preserves entries with large values in $\mathbf{P}$ while discarding those with negligible contribution. This is analogous to the router in mixture-of-experts~\cite{shazeer2017outrageously,lu2025moba}, where each token must be routed to a subset of important experts; here, each query block must be routed to a subset of high-contribution key--value pairs. Moreover, since modern accelerators are optimized for dense computation, unstructured sparsity rarely translates into real speedups. Block-sparse attention~\citep{dao2022flashattentionfastmemoryefficientexact} is necessary for hardware efficiency, where every $(B,B)$ block of $\mathbf{M}$ shares the same value. Thus, the sparse attention problem reduces to designing an accurate router that produces the block-level mask $\mathbf{M}$, consisting of $(\tfrac{L}{B}, \tfrac{L}{B})$ boolean entries, since all tokens within a $(B,B)$ tile share the same value. 

VSA~\cite{zhang2025faster} addresses this by applying mean pooling, producing coarse-branch representations $\mathbf{Q}_c, \mathbf{K}_c, \mathbf{V}_c\in \mathbb{R}^{(L/B) \times D}$, where each token corresponds to a 3D cube $(B_t, B_h, B_w)$. The coarse branch then performs dense attention $\mathbf{S}_c = \mathbf{Q}_c \mathbf{K}_c^\top/\sqrt{D}, 
\mathbf{P}_c = \text{Softmax}(\mathbf{S}_c), 
\mathbf{O}_c = \mathbf{P}_c \mathbf{V}_c
$. The router reuses $\mathbf{P}_c$ and applies a topK selection to identify the index of the highest-scoring key-value blocks for each query block:
$\mathbf{M} = \text{TopK}$($\mathbf{Q_c}\mathbf{K_c}^\top, \text{K}, \text{dim}=-1)$. A fine branch then uses $\mathbf{M}$ to perform token-level block-sparse attention.

Normally, the router topK, like those in mixture-of-experts language models~\cite{shazeer2017outrageously}, is learned with gradients propagated from the expert outputs. However, in sparse attention, while the coarse branch outputs $\mathbf{O}_c$ carry gradients during backpropagation, its router topK outputs $\mathbf{M}$ do not. Consequently, there is no gradient path from the fine branch back through the routing decision to update the coarse branch's routing capability. This naturally raises two questions: how does routing quality emerge and how can we further improve it? We consider two hypotheses. First, \emph{locality heuristics}: neighboring tokens tend to be similar, so mean pooling already yields informative cube features and the router need not receive trainable parameters nor gradients, in line with observations from MoBA~\cite{lu2025moba} and Quest~\cite{tang2024quest}. Second, \emph{auxiliary supervision}: the router shares parameters with the coarse branch, and the gradients from $\mathbf{O}_c$ benefit the router, despite the fact that the gradients do not come from $\mathbf{M}$, as in the case of NSA~\cite{yuan2025native}. In \S\ref{sec:exp_ablations}, we perform apples-to-apples ablations that isolate these factors and find that explicit gradient flow to the router is unnecessary. Based on this result, \methodnameshort~therefore focuses on improving the accuracy \emph{of gradient-free routing} by decoupling the pooling granularity of the router from the granularity of the block-sparse attention, so that each query block reliably identifies the most contributing key–value blocks.







\subsection{VSA2}
\label{sec:method:vsa2}

\begin{figure}[t]
    \centering
    \begin{minipage}[b]{0.49\textwidth}
        \centering
        \includegraphics[width=\linewidth]{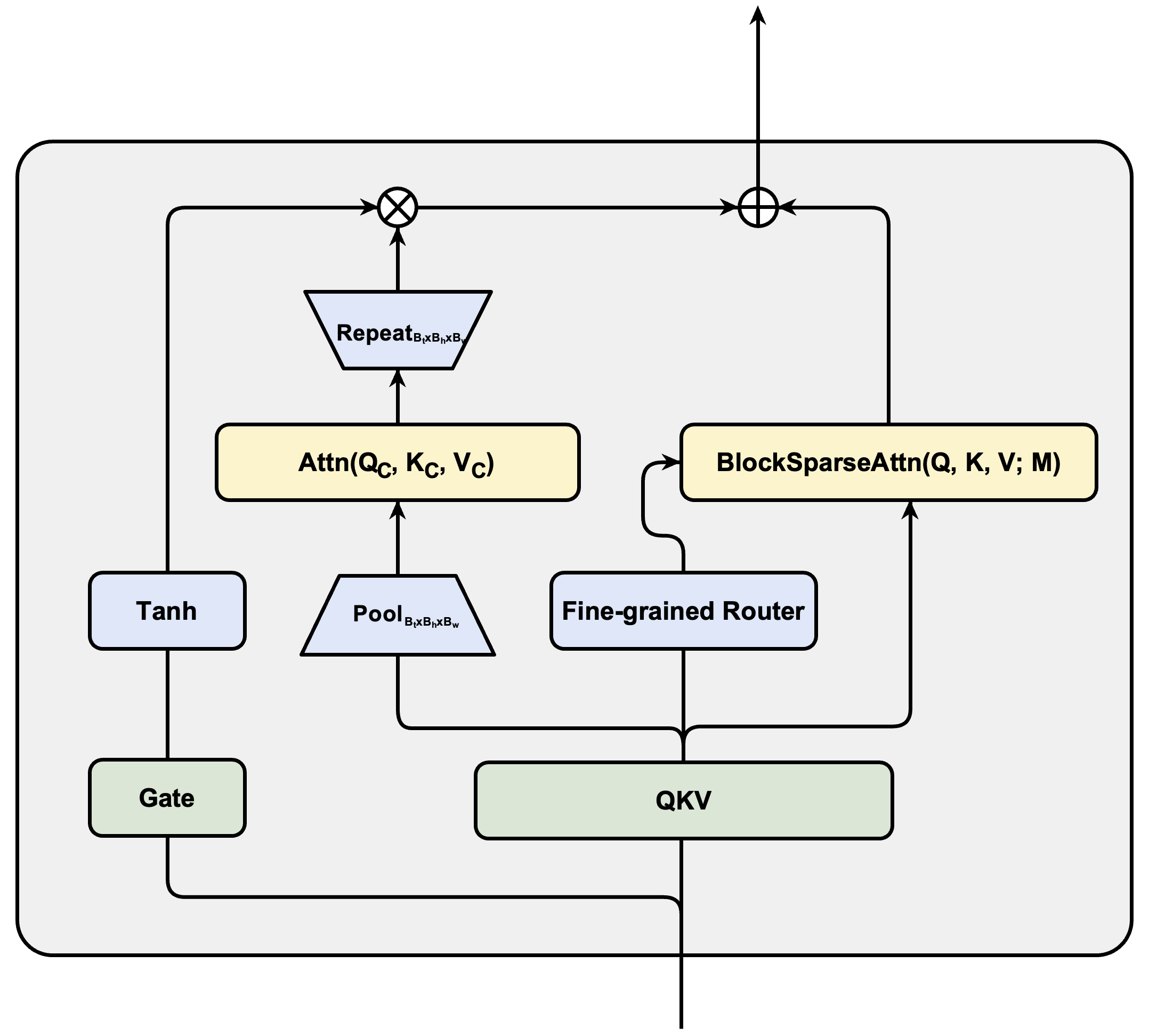}
    \end{minipage}\hfill%
    \begin{minipage}[b]{0.49\textwidth}
        \centering
        \includegraphics[width=\linewidth]{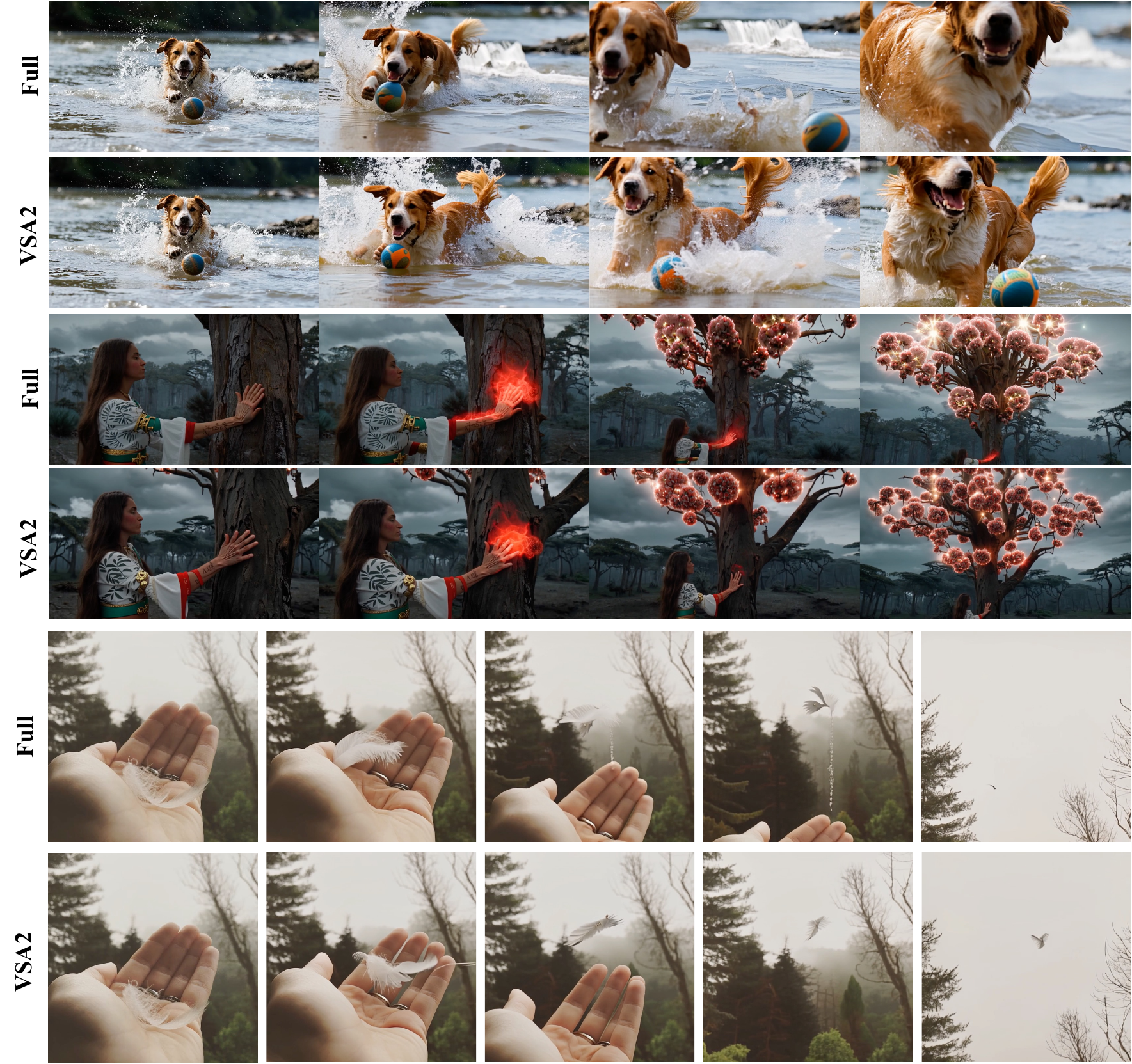}
    \end{minipage}
    \par
    \noindent
    \begin{minipage}[t]{0.49\textwidth}
        \vspace{0pt}
        \captionof{figure}{%
            \raggedright
            Architecture of \methodname, comprising a coarse branch,
            a fine branch, and a separate router with a
            pool-softmax-pool structure.
        }
        \label{fig:main_plot}
    \end{minipage}\hfill%
    \begin{minipage}[t]{0.49\textwidth}
        \vspace{0pt}
        \captionof{figure}{%
            \raggedright
            Qualitative comparison after pretraining + RL.
            The top and bottom examples use image-to-video generation;
            the middle uses text-to-video.
        }
        \label{fig:qualitative_examples}
    \end{minipage}
\end{figure}

Table~\ref{tab:vsa2_notation} summarizes our notation. As sketched in Figure~\ref{fig:main_plot}, \methodnameshort\ has three components—a \emph{coarse branch}, a \emph{fine branch}, and a \emph{router}—that work together to turn dense 3D attention into hardware-aligned block sparse attention while preserving key dependencies. 

\noindent \textbf{Permutation.}
We first permute the tokens into cube-major order:
$\mathbf{Q}, \mathbf{K}, \mathbf{V} \in \mathbb{R}^{THW \times D} = \mathbb{R}^{N_t N_h N_w \times B_t B_h B_w \times D} = \mathbb{R}^{N \times B \times D}$~\cite{zhang2025fastvideo, zhang2025faster}.
This permutation can be applied once at the beginning of the transformer with proper permutation of RoPE embeddings~\cite{su2024roformer}, as attention is the only operation that depends on token order.

\Needspace{9\baselineskip}
\noindent \textbf{Coarse Branch.}
The coarse branch captures high-level structure via cube-level full attention, which is the same as VSA:

\begin{align}
[\mathbf{Q}_c, \mathbf{K}_c, \mathbf{V}_c] &= \text{MeanPool}_B([\mathbf{Q}, \mathbf{K}, \mathbf{V}]) \in \mathbb{R}^{3N \times D}, \label{eq:compress}\\
\mathbf{P}_c &= \text{Softmax}\!\left(\mathbf{Q}_c \mathbf{K}_c^{\top}/\sqrt{D}\right) \in \mathbb{R}^{N \times N}, \label{eq:compress_score} \\
\hat{\mathbf{O}}_c &= \mathbf{P}_c \mathbf{V}_c
\in \mathbb{R}^{N \times D}.
\end{align}
where $\text{MeanPool}_B$ denotes mean pooling over the $(B_t, B_h, B_w)$ cube. In our early experiments, we also explored alternative pooling methods, including those similar to~\cite{yuan2025native} and attention-based pooling, but found them to be no better than simple mean pooling in video DiT. The coarse output is then broadcasted back to the original video resolution:

\begin{equation}
\mathbf{O}_c = \text{Repeat}_B(\hat{\mathbf{O}}_c) \in \mathbb{R}^{N \times B \times D}.
\end{equation}

\noindent \textbf{Router.}
In VSA, the coarse branch attention score $\mathbf{P}_c$ defines the cube-to-cube affinity score used to perform a Top-K selection and generate the block-sparse attention mask. In practice, we find the original pooling $B_t \times B_h \times B_w$ to be excessively large such that much information is lost in pooling. Instead, we decouple the router from the coarse branch and adopt a much smaller pooling size $R_t \times R_h \times R_w$:
\begin{align}
[\mathbf{Q}_r, \mathbf{K}_r] &= \text{MeanPool}_R([\mathbf{Q}, \mathbf{K}]) \in \mathbb{R}^{2N \times G \times D}, \label{eq:router_start} \\
\hat{\mathbf{P}}_r &= \text{Softmax}\!\left(\frac{\mathbf{Q}_r \mathbf{K}_r^{\top}}{\sqrt{D}}\right)
\in \mathbb{R}^{N \times N \times G \times G}.
\end{align}
After softmax, we aggregate the router score to block scores using another pooling operator:
\begin{equation}
\mathbf{P}_r = \text{MeanPool}_{G \times G}(\hat{\mathbf{P}}_r)
\in \mathbb{R}^{N \times N} \label{eq:router_end} .
\end{equation}
Based on $\mathbf{P}_r$, we select the topK key blocks at the sequence level rather than per-token topK:
\begin{align}
\mathbf{P}_r &\in \mathbb{R}^{N \times N}
\;\;\longrightarrow\;\;
\mathbf{P}_r \in \mathbb{R}^{N N}, \\
\mathbf{M} &= \text{TopK}_{N K}(\mathbf{P}_r),
\end{align}
which yields $N K$ selected blocks per sequence.
This design maintains a consistent compute budget while allowing each query to attend to different numbers of key-value blocks.

The intuition behind the pool-softmax-pool operation in Eq.~\eqref{eq:router_start}–\eqref{eq:router_end} is simple. In the compute-unbounded case, an oracle router would first compute token-level attention and only then aggregate to block-level affinities:
\begin{align}
\mathbf{P} = \text{Softmax}\!\left(\frac{\mathbf{Q} \mathbf{K}^{\top}}{\sqrt{D}}\right)\in \mathbb{R}^{N \times N \times B \times B}, \\
\mathbf{P}_{oracle} = \text{MeanPool}_{B \times B}(\hat{\mathbf{P}})
\in \mathbb{R}^{N \times N}.
\end{align}
Equations~\eqref{eq:compress}–\eqref{eq:compress_score} approximate $\mathbf{P}_{oracle}$ by moving the pooling step: instead of the full attention matrix $\hat{\mathbf{P}}$, it pools $\mathbf{Q}$ and $\mathbf{K}$~\emph{before} applying the dot product and softmax.
This reordering would be exact if attention were linear, but because softmax is nonlinear, averaging $\mathbf{Q}$ and $\mathbf{K}$ is not equivalent to averaging $\hat{\mathbf{P}}$. As a result, the method provides an approximation rather than an identity. 
Its accuracy depends on a locality assumption: tokens within a cube $B_t \times B_h \times B_w$ are similar enough that pre-averaging them introduces minimal distortion.

Our fine-grained router, shown in Figure~\ref{fig:router_comparison} and defined by equations \eqref{eq:router_start}–\eqref{eq:router_end}, occupies a middle ground between the oracle scores
$\mathbf{P}_{\mathrm{oracle}}$ and the coarse-branch approximation $\mathbf{P}_c$. The key idea is to decouple the router’s pooling size $R$ from the block size $B$ used in block sparse attention.
We first apply a much \emph{smaller} pooling $R=R_t\times R_h\times R_w$ to $\mathbf{Q}$ and $\mathbf{K}$,
which reduces sequence length and router FLOPs while preserving substantially more intra-cube detail than $B$-pooling.
We then compute the router's attention $\hat{\mathbf{P}}_r$ and aggregate it with a $G{\times}G$ operator to obtain
block-level scores $\mathbf{P}_r\in\mathbb{R}^{N\times N}$.
This design preserves the fine-grained heterogeneity that the coarse branch overlooks, yet keeps routing cost negligible. 

\noindent \textbf{Fine Branch.}
The fine branch only includes a block-sparse attention computed as:
\begin{equation}
\mathbf{O}_f = \text{BlockSparseAttn}(\mathbf{Q}, \mathbf{K}, \mathbf{V}; \mathbf{M})
\in \mathbb{R}^{N \times B \times D},\label{eq:block_sparse}
\end{equation}
where BlockSparseAttn denotes block-sparse attention restricted to the selected block pairs in $\mathbf{M}$. This generalizes to cross attention for text conditioning as in MMDiT~\cite{esser2024scaling}, as Eq.~\eqref{eq:block_sparse} extends to include full video-to-text and text-to-video attention components.

\noindent \textbf{Gated Merge.}
The coarse and fine branch outputs are combined through a lightweight gating mechanism:
\begin{align}
\mathbf{G} &= \tanh\!\big(\mathbf{X}\mathbf{W}_g)\label{eq:gate_linear},\\
\mathbf{O} &= \mathbf{G}\odot \mathbf{O}_c + \mathbf{O}_f.
\end{align}
Here, $\mathbf{X}\in\mathbb{R}^{L\times D}$ are the hidden states of the transformer layer, and $\mathbf{W}_g\in\mathbb{R}^{D\times 1}$ is a single-channel projection. This produces a gate per-token per-head $\mathbf{G}\in\mathbb{R}^{L\times 1}$ \citep{qiu2025gated}, which is broadcasted to $\mathbb{R}^{L\times D}$ to modulate $\mathbf{O}_c$; $\odot$ denotes element-wise multiplication.

\noindent \textbf{Kernel Implementation.}
For non-divisible $(T,H,W)$, we pad each dimension to the nearest multiple of $(B_t,B_h,B_w)$ and ensure the attention to padding tokens is masked out in our kernels. Block-sparse attention is implemented in ThunderKittens~\cite{spector2024thunderkittens}, following \citep{zhang2025faster}. To avoid materializing $\hat{\mathbf{P}}_r$, we fuse the GEMM, softmax, and $(G{\times}G)$-pooling into a single kernel implemented with CuTe DSL. The fused kernel computes $\mathbf{Q}_r\mathbf{K}_r^{\top}$ twice (Figure~\ref{fig:router_kernel_plot}): a first pass to accumulate log-sum-exp (LSE) statistics, and a second to apply softmax, perform score pooling, and write back. Although this introduces one additional compute pass, it substantially cuts I/O and yields a clear speedup over the unfused baseline.

\subsection{Improved Training Recipe}
\label{sec:method:training}

\begin{table*}[t]
\centering
\setlength{\tabcolsep}{2pt} 
\resizebox{\textwidth}{!}{%
\begin{tabular}{llcccc|ccccc}
\toprule
\multirow{2}{*}{\textbf{Exp}} & \multirow{2}{*}{\textbf{Stage}} &
\multicolumn{4}{c|}{\textbf{Configuration}} & \multicolumn{3}{c}{\textbf{T2V}} & \multicolumn{2}{c}{\textbf{I2V}} \\
\cmidrule(lr){3-6} \cmidrule(lr){7-9} \cmidrule(lr){10-11}
 & & Train TopK & Infer TopK & Attn Sparsity & E2E Speedup & Motion & Following & Aesthetics & Motion & Following \\
\midrule
1 & 480p & 64 & 64 & 0.90 & 2.09$\times$ & -1.34\% & 1.34\% & -0.67\% & 8.72\% & 1.34\% \\
2 & 480p & 64 & 128 & 0.82 & 1.92$\times$ & 22.1\% & -3.36\% & -4.7\% & 6.71\% & -8.72\% \\
3 & 480p & 32 & 128 & 0.82 & 1.92$\times$ & 7.38\% & -15.4\% & -4.7\% & 11.4\% & -9.4\% \\
4 & 480p & 32 & 64 & 0.90 & 2.09$\times$ & 12.1\% & -10.1\% & -7.38\% & 4.03\% & -4.03\% \\ 
5 & 480p RL & 64 & 64 & 0.90 & 2.09$\times$ & -4.03\% & 1.34\% & -2.01\% & 8.05\% & -4.7\% \\
6 & 480p RL & 64 + 128 & 128 & 0.82 & 1.92$\times$ & 7.38\% & -2.01\% & 0.00\% & 8.05\% & -2.69\% \\
7 & 720p & 64 & 64 & 0.95 & 4.62$\times$ & 6.71\% & -2.01\% & 1.34\% & 0.67\% & -1.34\% \\
\bottomrule
\end{tabular}%
}
\caption{Human evaluation results of \methodnameshort~against full attention across various stages of video DiT training. A positive score means the model trained with \methodnameshort~has better quality than its full attention counterpart at that stage.}
\label{tab:main_t2v_i2v_human}
\end{table*}

Instead of training from scratch, we develop a more efficient strategy, which we term \upcyclingname~for our main experiments in \S\ref{sec:exp_main_results}. This approach initializes the model from a pre-trained 256p video checkpoint that uses full attention and then introduces \methodnameshort~for subsequent training on 480p and 720p resolutions, and reinforcement learning. The transition is seamless: the gating weights $\mathbf{W_g}$ in Eq.~\eqref{eq:gate_linear}, which are the only newly added parameters, are initialized to zero, ensuring the model's output contains only $\mathbf{O_f}$ initially. This strategy offers two significant advantages. First, it allows us to leverage existing text-to-image models that are trained with full attention, which is a common practice for state-of-the-art video model training pipelines~\cite{wan2025wan, zheng2024open}. Second, it is flexible, as the benefits of sparse attention are minimal at the early, low-resolution stage of video training, where sequence lengths are still short. As demonstrated in \S\ref{sec:exp_main_results}, a model trained through \upcyclingname~even achieves better quality than its full-attention counterpart. Furthermore, we find that reducing the sparsity level during inference—a schedule we call \curriculumname~(train-hard/test-easy)—improves motion quality, even surpassing the performance of the original full-attention model.

\section{Experiments}

\begin{figure}[t]
    \centering
    \includegraphics[width=\linewidth]{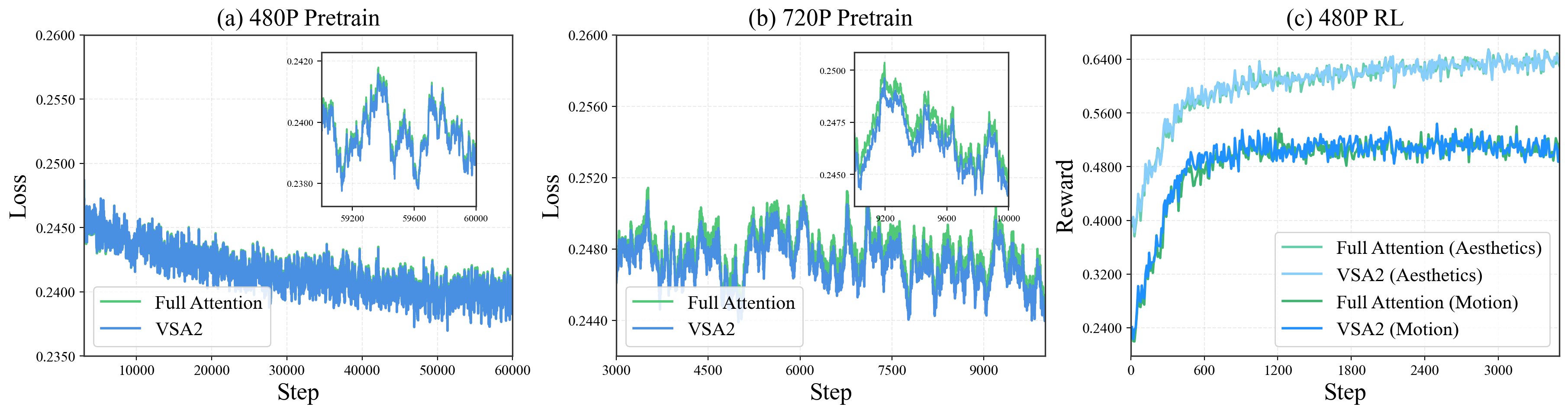}
    \caption{Training loss and reward scores of \methodnameshort~vs. full attention across various stages of video DiT training. With a sparsity ratio of 90\% to 95\%, \methodnameshort~shows lower loss and similar reward scores compared to full attention.}
    \label{fig:main_loss}
\end{figure}

\subsection{Setup}


\noindent \textbf{Model Training.} We largely follow the architecture of MMDiTs~\cite{esser2024scaling}. Training uses flow matching~\cite{lipmanflow, liu2022flow} with a velocity prediction objective and joint text-to-video and image-to-video supervision on a video dataset. Timestep sampling follows a logit-normal distribution with resolution-aware shifting~\cite{esser2024scaling}. We employ FSDP~\cite{zhao2023pytorch}, sequence parallelism~\cite{jacobs2023deepspeed}, activation recomputation~\cite{chen2016training}, and \texttt{torch.compile}~\cite{ansel2024pytorch} for scalable multi-node training. Training proceeds in three stages: 480p and 720p pretraining, followed by an RL stage~\cite{xu2023imagereward} initialized from 480p checkpoints. Additional details on the RL stage are provided in the supplementary. Across all stages, training clips span 5–12 seconds with diverse aspect ratios.

\begin{figure}[t]
    \centering
    \begin{minipage}[t]{0.56\textwidth}
        \vspace{0pt}
        \centering
        \includegraphics[width=\linewidth]{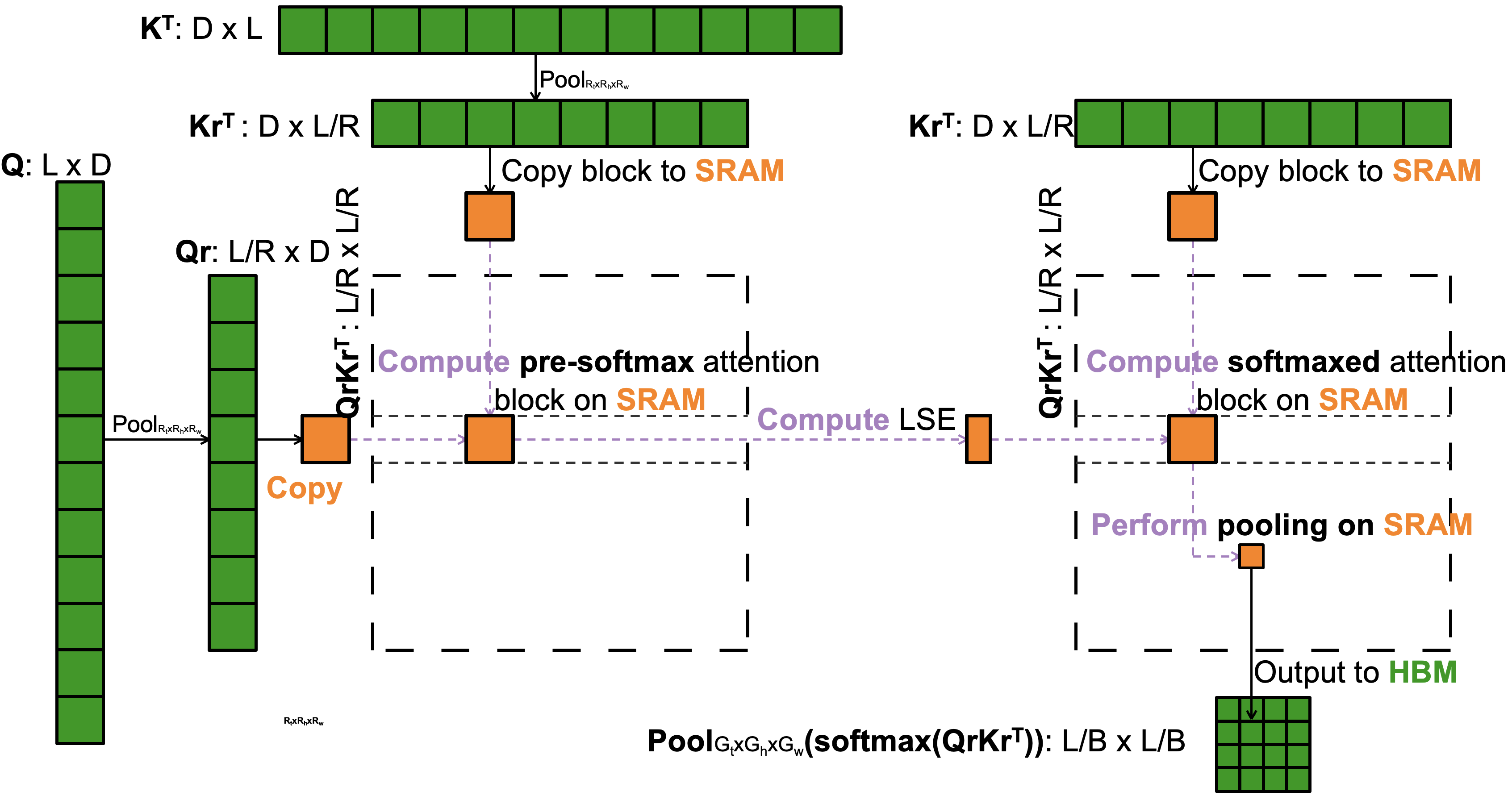}
        \captionof{figure}{Kernel implementation of \methodnameshort's router,
        which consists of two GEMM passes. The first pass computes
        the LSE for softmax and the second pass calculates the
        softmax score with on-chip pooling.}
        \label{fig:router_kernel_plot}
    \end{minipage}\hfill
    \begin{minipage}[t]{0.42\textwidth}
        \vspace{0pt}
        \centering
        \includegraphics[width=\linewidth]{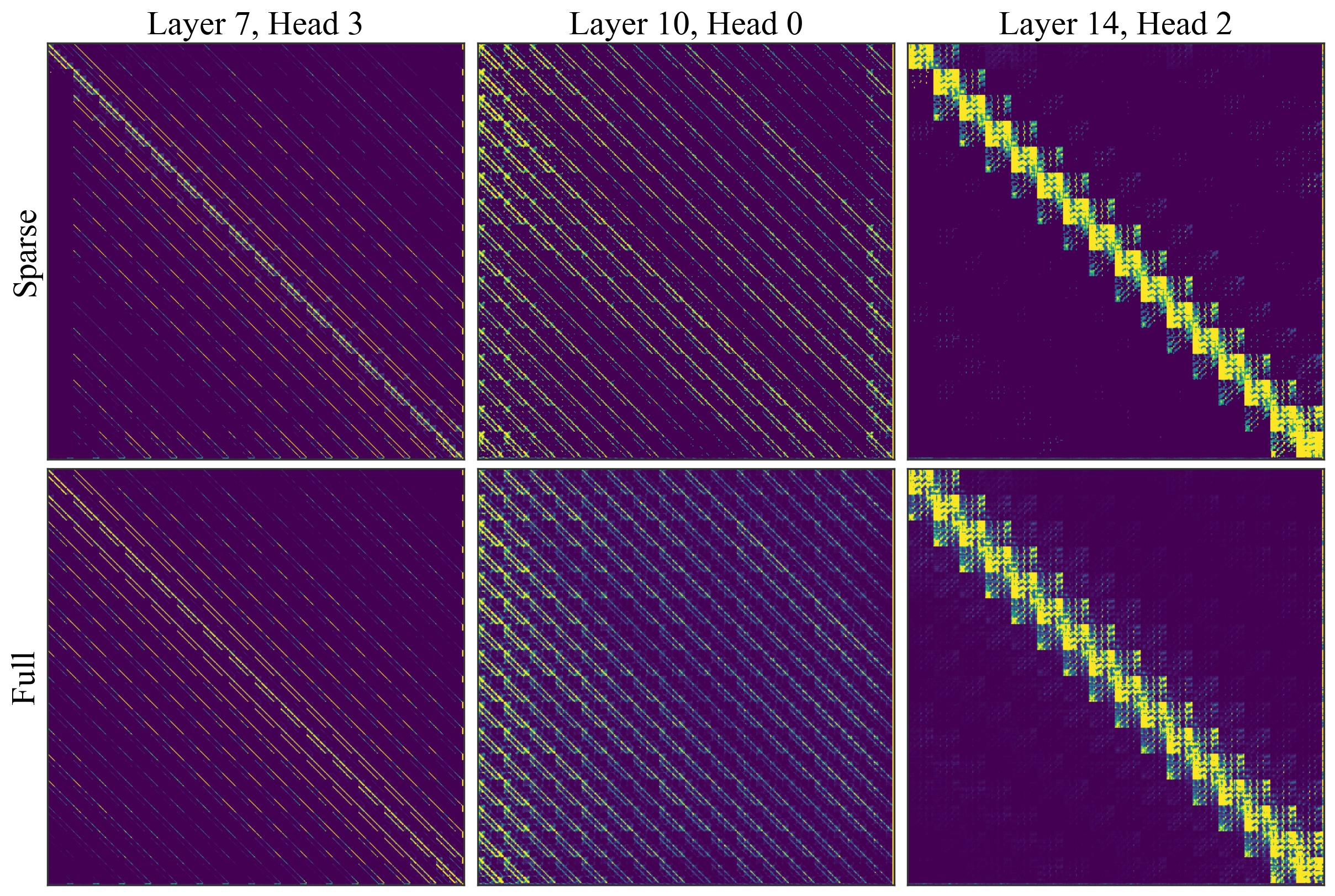}
        \captionof{figure}{Attention visualization of \methodnameshort~vs.
        full attention. Their attention patterns are highly similar
        even after pretraining.}
        \label{fig:attn_visualization}
    \end{minipage}
\end{figure}

\noindent \textbf{Baselines.}
In the main results, we compare \methodnameshort~against full attention baselines trained under exactly the same hyperparameters, data, and iterations. Architecturally, \methodnameshort~can be understood as VSA~\cite{zhang2025vsa} with a fine-grained router and per-sequence topK, and we ablate those changes in \S\ref{sec:exp_ablations}. We also compare \methodnameshort~with other choices of efficient attention designs, including key architectures of NSA~\cite{yuan2025native}. To ensure a fair comparison, all models are configured to have a similar parameter count.


\noindent \textbf{Evaluation Metric.} We primarily use flow matching loss as the evaluation metric, which correlates well with human preference~\cite{polyak2025moviegencastmedia}. We additionally conduct human evaluation on 149 curated prompts comparing \methodnameshort~and full attention, with each pair rated as \textit{good}, \textit{same}, or \textit{bad}. The final score is computed as $(\#\text{good} - \#\text{bad}) / 149$ and reported in Table~\ref{tab:main_t2v_i2v_human}. The prompts are intentionally challenging to better distinguish model performance. For evaluation, we use EMA checkpoints to generate 10-second videos at resolutions of $480\times864$ (99k tokens) and $720\times1280$ (220k tokens).

\subsection{Main Results}
\label{sec:exp_main_results}

As demonstrated in Table~\ref{tab:main_t2v_i2v_human} Exps.~1, 5, and 7 and Figure~\ref{fig:main_loss}, \methodnameshort~performs comparably to full attention in all training stages, including 480p, 720p, and RL. This strong performance is consistent across human evaluations, flow-matching loss, and RL rewards. A key finding is that as the resolution increases from 480p to 720p, \methodnameshort~does not require a higher topK value to achieve a lower loss than the full-attention baseline. At 720p resolution, this efficiency results in a remarkable 95\% sparsity and a 4.62$\times$ increase in end-to-end inference speed.

Comparing Exps.~1--4 in Table~\ref{tab:main_t2v_i2v_human}, we observe a consistent trend: employing a higher topK value during inference than during training—a strategy we term \curriculumname—significantly improves motion quality compared to the full-attention baseline. In particular, in Exp.~2, human raters judged 22.1\% of \methodnameshort~text-to-video samples to exhibit better motion than those from the full attention model. A model trained with top32 and tested with top64 (Exp.~4) achieves better motion scores than one both trained and tested with top64 (Exp.~1). This improvement, however, is accompanied by a trade-off in prompt following and aesthetic quality.


We hypothesize that the improved motion quality stems from a regularization effect similar to structured dropout~\cite{ghiasi2018dropblock}, while reduced prompt adherence arises from a training–inference mismatch. In MMDiT, video and text tokens share the same self-attention space; increasing the number of attended video tokens at inference effectively reduces attention paid to text tokens. A lightweight fine-tuning stage with a higher topK can mitigate this issue. Indeed, in Exp.~6 of Table~\ref{tab:main_t2v_i2v_human}, RL training with top128 initialized from a top64 checkpoint restores prompt following and aesthetic quality while preserving superior motion quality over the full-attention baseline. Figure~\ref{fig:qualitative_examples} shows human-evaluation samples from Exp.~6, with additional results in the supplementary material. RL at a higher TopK allows the model to re-adapt attention allocation under denser contexts, effectively bridging the gap between sparse pretraining and less sparse inference. We further show in the supplementary that \methodnameshort, trained on 5--12\,s videos, can directly generate 30\,s outputs without noticeable degradation, providing early evidence that VSA2 scales toward the minute-long regime.


To further analyze model behavior, we visualize and compare attention maps from \methodnameshort~and a full-attention model in Figure~\ref{fig:attn_visualization}. Unlike prior work that profiles full-attention models at inference time~\cite{xi2025sparsevideo,zhang2025sparge,zhang2025fast}, we compare two distinct checkpoints from Exp.~1 in Table~\ref{tab:main_t2v_i2v_human}, one trained with \methodnameshort~and the other with full attention. As shown in Figure~\ref{fig:attn_visualization}, the attention maps remain remarkably similar even after 60K training steps, providing strong evidence that \methodnameshort~preserves training dynamics comparable to full attention during pretraining.

\subsection{Ablation Studies}
\label{sec:exp_ablations}

\begin{figure}[t]
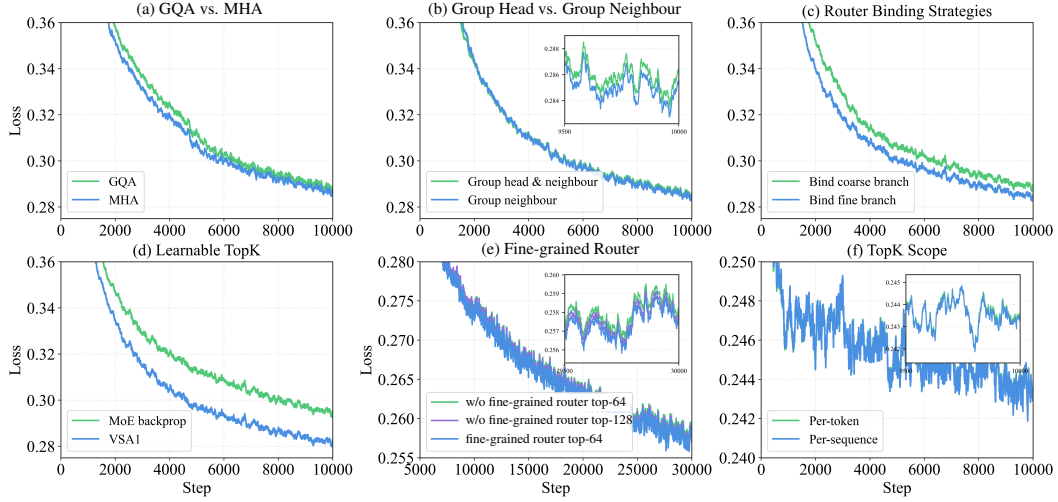

    \centering
    \ablationillustration
    \caption{Ablation studies of various design choices in video sparse attention. All models are trained from scratch except those in plot (f), which are initialized from a 256p video checkpoint.}
    \label{fig:ablation_loss}
\end{figure}

In this section, we present a series of ablation studies to justify the key design choices of \methodnameshort. We begin by analyzing the token grouping and router mechanisms, which represent the primary architectural departures from NSA~\cite{yuan2025native}. Subsequently, we ablate the specific enhancements that differentiate \methodnameshort~from VSA~\cite{zhang2025faster}, providing a comprehensive validation of our attention design.


\noindent \textbf{Group Head vs. Group Neighbour.}
Sparse attention in language models, such as NSA~\cite{yuan2025native}, is based on grouped-query attention (GQA). NSA adopts a ``group head'' approach, where query heads within the same group share the same sparsity pattern. Conversely, \methodnameshort~utilizes Multi-Head Attention (MHA) and adopts a ``group neighbour'' strategy, where spatially and temporally neighboring query tokens attend to a common set of KV tokens. To validate our architectural choice of MHA+group neighbour over GQA+group head, we conducted two ablation studies. First, we compare GQA vs. MHA under a full attention setting. As shown in Figure~\ref{fig:ablation_loss}(a), MHA achieves a significantly lower training loss than GQA (with 4 groups), suggesting that it is inherently better suited for video generation. Second, we investigate the grouping strategy itself. Figure~\ref{fig:ablation_loss}(b) demonstrates that even within a GQA framework, the ``group neighbour'' approach outperforms a hybrid ``group head \& neighbour'' strategy. For this experiment, the ``group neighbour'' method used larger spatiotemporal cubes, while the hybrid method combined smaller cubes with head grouping. Both configurations maintained the same effective group size.
 The better performance of the pure ``group neighbour'' configuration confirms that it is the more effective strategy for video models.

\noindent \textbf{Learnable Router vs. Gradient-free Router.}
As discussed in \S\ref{sec:background}, VSA/NSA~\cite{yuan2025native} use a learnable router that shares parameters with the coarse branch and receives gradients from $\mathbf{O_c}$. We conducted ablation studies, presented in Figures~\ref{fig:ablation_loss}(c) and (d), to determine if such a learnable router is really necessary for video models. First, we decouple the QKV projection layers for the coarse and fine attention branches. We compared two configurations: (1) share the router QKV with the coarse branch's QKV, where the router parameters receive gradients from $\mathbf{O_c}$, and (2) share the router QKV with the fine branch's QKV, which is analogous to a gradient-free router design in MoBA~\cite{lu2025moba}. The results in Figure~\ref{fig:ablation_loss}(c) demonstrate that binding the router to the fine branch achieves a lower training loss, indicating that the router does not need to receive gradients from $\mathbf{O_c}$. Next, we investigate an alternative learnable router design inspired by mixture-of-experts in FFN. In this experiment, the coarse branch, the router and the fine branch were assigned an independent set of QKV parameters. The output of the fine branch was re-weighted by the router's output logits, allowing the router logits to receive direct gradients based on the importance it assigns to each block. As shown in Figure~\ref{fig:ablation_loss}(d), this MoE-style backpropagation approach did not outperform VSA. We thus conclude that a complex, learnable router is not necessary for video diffusion models. A simpler, effectively gradient-free router is sufficient and yields better performance.

\noindent \textbf{Gradient-Free Router Design.}
To optimize the gradient-free router design, we introduce a \textit{fine-grained router}. The efficacy of this design is demonstrated in Figure~\ref{fig:ablation_loss}(e), which shows that models incorporating the fine-grained router achieve a significantly lower training loss compared to the non-fine-grained router baseline. Notably, the fine-grained router with a top64 outperforms the top128 baseline router. Furthermore, we investigated the optimal scope for the topK selection strategy. As illustrated in Figure~\ref{fig:ablation_loss}(f), applying topK selection on a \textit{per-sequence} basis yields a lower loss than the conventional \textit{per-token} approach.
We also tried some top-p-based methods during preliminary studies, but these did not yield positive results. These ablation studies, which favor the fine-grained router and per-sequence topK, collectively inform the final architectural design of \methodnameshort.

\subsection{Attention Speed}
\label{sec:exp_attn_speed}

\begin{wrapfigure}{r}{0.44\textwidth}
    \vspace{-1em}
    \centering
    \includegraphics[width=0.42\textwidth]{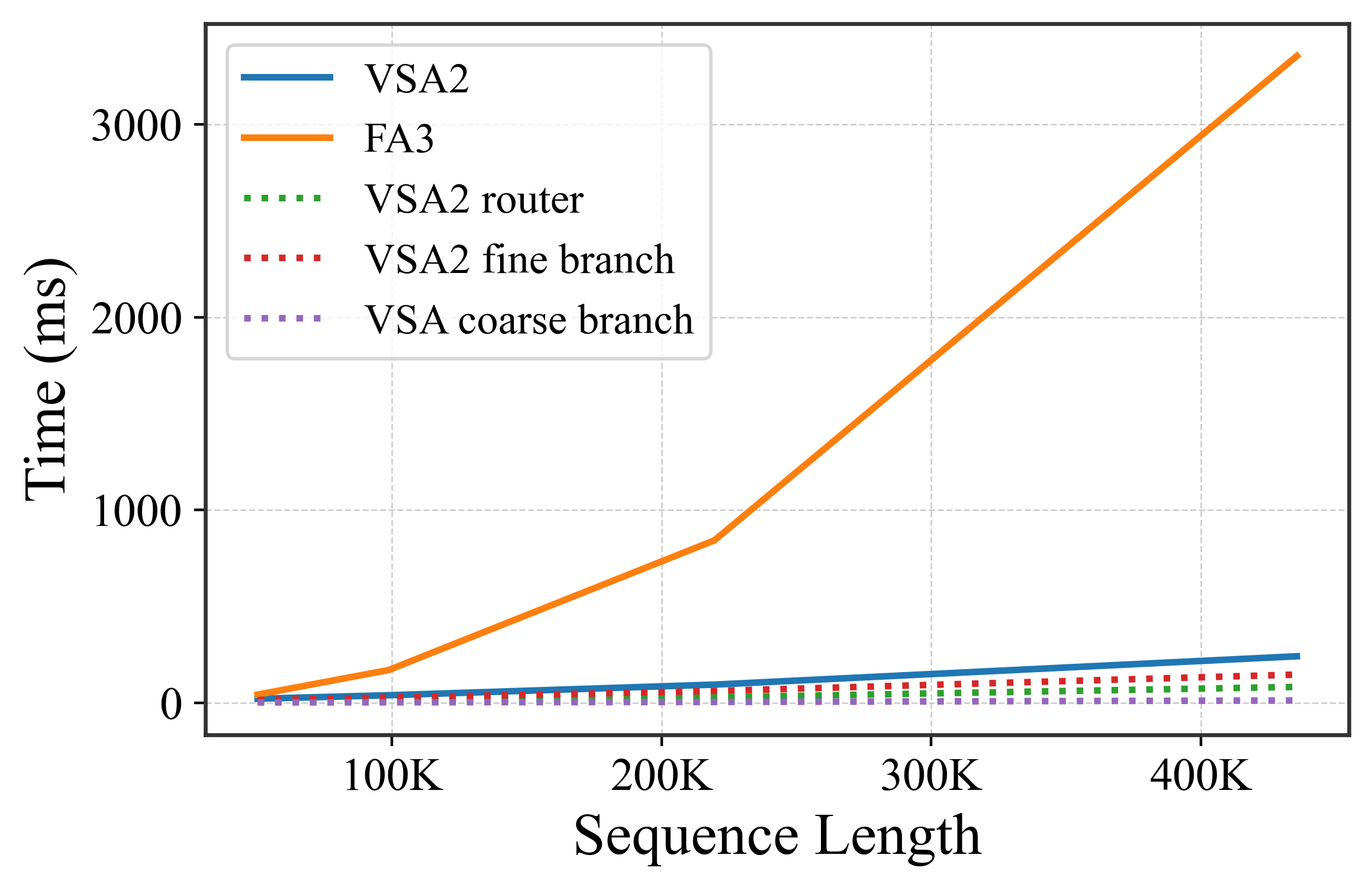}
    \vspace{-1em}
    \caption{Speed comparison on a single H800 GPU with batch size 1, number of heads 20, head dimension 128, and top64. }
    \vspace{-1em}
    \label{fig:kernel_speed}
\end{wrapfigure}

Table~\ref{tab:main_t2v_i2v_human} presents the model-level speedup of \methodnameshort, and Figure~\ref{fig:kernel_speed} compares its attention-level performance against FlashAttention-3~\cite{shah2024flashattention}. For this comparison, we report the total time of all operations in Figure~\ref{fig:main_plot}, excluding the QKV projections. We also provide a detailed runtime breakdown for the coarse branch, fine branch, and router components. \methodnameshort~achieves an $8.9\times$ acceleration at a sequence length of 220K (corresponding to a 720p--10s video). The fine-grained router accounts for 22\% of the total attention runtime at 220K and 30\% at 436K. Considering that the fine-grained router reduces the fine branch computation by half (Figure~\ref{fig:ablation_loss}~(e)), its additional cost is well justified.

\section{Related Work}

\noindent \textbf{Video DiTs.}
Modern video DiTs follow a multi-stage training recipe. First, large-scale pretraining adopts a progressive curriculum: training starts on images, then advances to short low-resolution clips, and finally scales to long high-resolution videos~\cite{lin2024open, zheng2024open, kong2024hunyuanvideo, wan2025wan, gao2025seedance}. Second, \emph{supervised fine-tuning} (SFT) on curated text--video pairs improves motion, instruction-following, and style control. \emph{Third}, \emph{reinforcement learning} tailors generation to human feedback with diffusion-adapted RL objectives, which report consistent gains in semantic alignment and temporal coherence~\cite{wu2025rewarddance, xue2025dancegrpo,shen2025directly, liu2025videodpo}. Finally, \emph{step distillation} reduces the diffusion steps~\cite{lu2025hyper,yin2024improved, yin2024onestep}. In contrast to LLMs, where most pretraining FLOPs occur on short contexts, video DiTs expend the bulk of compute on high-resolution training, making quadratic attention a bottleneck at both pretraining and inference. 

\noindent \textbf{DiT Inference Acceleration.}
The iterative denoising process in DiTs is computationally intensive. Step distillation methods, such as progressive distillation~\cite{salimans2022progressive} and consistency distillation~\cite{song2023consistency,song2023improved,wang2024phased}, train a student model to replicate a teacher's output in fewer steps~\cite{salimans2022progressive}. More recent data-free methods align student and teacher models by matching intermediate distributions~\cite{yin2024onestep,yin2024improved}.
In addition to step reduction, caching methods exploit temporal redundancy by reusing intermediate activations from previous denoising steps, thus avoiding redundant computations~\cite{ma2024deepcache, ma2024learning, lv2024fastercache, kahatapitiya2025adaptive}. Lastly, post-training quantization has enabled W8A8 or even W4A4 precision for DiTs with minimal quality loss~\cite{wan2025wan,li2024svdquant,mehta2025tensorrt}, and attention-specific 8-bit/4-bit quantization offers plug-and-play acceleration~\cite{zhang2024sageattention,zhang2024sageattention2,zhang2025sageattention3}. Sparse attention, as the focus of this paper, is largely orthogonal to those methods~\cite{fastvideo2025posttraining}.

\noindent \textbf{Sparse Attention.}
Sparse attention mitigates the quadratic complexity of self-attention by restricting the computation to a smaller, more relevant subset of critical tokens. In LLMs, sparse attention has been extensively studied, with methods ranging from fixed sparsity patterns that target specific token arrangements~\cite{yang2408post, xiao2023efficient, jiang2024minference} to input-adaptive approaches that dynamically identify salient regions for attention computation~\citep{tang2024quest,gao2024seerattention,zhang2023h2o,lu2025moba,yuan2025native}.
Adapting these techniques to video introduces unique spatio-temporal challenges. Many recent training-free methods for video models introduce sparsity by exploiting locality or pre-defined temporal patterns~\cite{yuan2024ditfastattn,zhang2025fastvideo,ding2025efficient,xu2025xattention,li2025radial,xi2025sparsevideo}. Others generate sparse masks dynamically during inference~\cite{xi2025sparsevideo,zhao2025paroattention,cai2025mixture}. 
Although pioneering works like DSV~\cite{tan2025dsv} and VSA~\cite{zhang2025faster} have explored integrating sparse attention directly into the pre-training phase, their studies have been confined to smaller-scale experiments, with evaluation relying primarily on the loss metric. To our knowledge, \methodnameshort~is the first sparse attention mechanism validated at various stages of the DiT development cycle, including pre-training, subsequent RL with human preference, and inference.



\bibliography{main}

\begin{thebibliography}{69}
\providecommand{\natexlab}[1]{#1}
\providecommand{\url}[1]{\texttt{#1}}
\expandafter\ifx\csname urlstyle\endcsname\relax
  \providecommand{\doi}[1]{doi: #1}\else
  \providecommand{\doi}{doi: \begingroup \urlstyle{rm}\Url}\fi

\bibitem[Ansel et~al.(2024)Ansel, Yang, He, Gimelshein, Jain, Voznesensky, Bao, Bell, Berard, Burovski, et~al.]{ansel2024pytorch}
Jason Ansel, Edward Yang, Horace He, Natalia Gimelshein, Animesh Jain, Michael Voznesensky, Bin Bao, Peter Bell, David Berard, Evgeni Burovski, et~al.
\newblock Pytorch 2: Faster machine learning through dynamic python bytecode transformation and graph compilation.
\newblock In \emph{Proceedings of the 29th ACM International Conference on Architectural Support for Programming Languages and Operating Systems, Volume 2}, pp.\  929--947, 2024.

\bibitem[Ball et~al.(2025)Ball, Bauer, Belletti, Brownfield, Ephrat, Fruchter, Gupta, Holsheimer, Holynski, Hron, Kaplanis, Limont, McGill, Oliveira, Parker-Holder, Perbet, Scully, Shar, Spencer, Tov, Villegas, Wang, Yung, Baetu, Berbel, Bridson, Bruce, Buttimore, Chakera, Chandra, Collins, Cullum, Damoc, Dasagi, Gazeau, Gbadamosi, Han, Hirst, Kachra, Kerley, Kjems, Knoepfel, Koriakin, Lo, Lu, Mehring, Moufarek, Nandwani, Oliveira, Pardo, Park, Pierson, Poole, Ran, Salimans, Sanchez, Saprykin, Shen, Sidhwani, Smith, Stanton, Tomlinson, Vijaykumar, Wang, Wingfield, Wong, Xu, Yew, Young, Zubov, Eck, Erhan, Kavukcuoglu, Hassabis, Ghahramani, Hadsell, van~den Oord, Mosseri, Bolton, Singh, and Rockt{\"a}schel]{genie3}
Philip~J. Ball, Jakob Bauer, Frank Belletti, Bethanie Brownfield, Ariel Ephrat, Shlomi Fruchter, Agrim Gupta, Kristian Holsheimer, Aleksander Holynski, Jiri Hron, Christos Kaplanis, Marjorie Limont, Matt McGill, Yanko Oliveira, Jack Parker-Holder, Frank Perbet, Guy Scully, Jeremy Shar, Stephen Spencer, Omer Tov, Ruben Villegas, Emma Wang, Jessica Yung, Cip Baetu, Jordi Berbel, David Bridson, Jake Bruce, Gavin Buttimore, Sarah Chakera, Bilva Chandra, Paul Collins, Alex Cullum, Bogdan Damoc, Vibha Dasagi, Maxime Gazeau, Charles Gbadamosi, Woohyun Han, Ed~Hirst, Ashyana Kachra, Lucie Kerley, Kristian Kjems, Eva Knoepfel, Vika Koriakin, Jessica Lo, Cong Lu, Zeb Mehring, Alex Moufarek, Henna Nandwani, Valeria Oliveira, Fabio Pardo, Jane Park, Andrew Pierson, Ben Poole, Helen Ran, Tim Salimans, Manuel Sanchez, Igor Saprykin, Amy Shen, Sailesh Sidhwani, Duncan Smith, Joe Stanton, Hamish Tomlinson, Dimple Vijaykumar, Luyu Wang, Piers Wingfield, Nat Wong, Keyang Xu, Christopher Yew, Nick Young, Vadim Zubov, Douglas
  Eck, Dumitru Erhan, Koray Kavukcuoglu, Demis Hassabis, Zoubin Ghahramani, Raia Hadsell, A{\"a}ron van~den Oord, Inbar Mosseri, Adrian Bolton, Satinder Singh, and Tim Rockt{\"a}schel.
\newblock Genie 3: A new frontier for world models.
\newblock 2025.

\bibitem[Cai et~al.(2025)Cai, Yang, Zhang, Guo, Xiao, Yang, Xu, Yang, Yuille, Guibas, et~al.]{cai2025mixture}
Shengqu Cai, Ceyuan Yang, Lvmin Zhang, Yuwei Guo, Junfei Xiao, Ziyan Yang, Yinghao Xu, Zhenheng Yang, Alan Yuille, Leonidas Guibas, et~al.
\newblock Mixture of contexts for long video generation.
\newblock \emph{arXiv preprint arXiv:2508.21058}, 2025.

\bibitem[Chen et~al.(2016)Chen, Xu, Zhang, and Guestrin]{chen2016training}
Tianqi Chen, Bing Xu, Chiyuan Zhang, and Carlos Guestrin.
\newblock Training deep nets with sublinear memory cost.
\newblock \emph{arXiv preprint arXiv:1604.06174}, 2016.

\bibitem[Dao et~al.(2022)Dao, Fu, Ermon, Rudra, and Ré]{dao2022flashattentionfastmemoryefficientexact}
Tri Dao, Daniel~Y. Fu, Stefano Ermon, Atri Rudra, and Christopher Ré.
\newblock Flashattention: Fast and memory-efficient exact attention with io-awareness, 2022.
\newblock URL \url{https://arxiv.org/abs/2205.14135}.

\bibitem[Ding et~al.(2025)Ding, Li, Su, Zhang, Deng, Stoica, and Zhang]{ding2025efficient}
Hangliang Ding, Dacheng Li, Runlong Su, Peiyuan Zhang, Zhijie Deng, Ion Stoica, and Hao Zhang.
\newblock Efficient-vdit: Efficient video diffusion transformers with attention tile.
\newblock \emph{arXiv preprint arXiv:2502.06155}, 2025.

\bibitem[Esser et~al.(2024)Esser, Kulal, Blattmann, Entezari, M{\"u}ller, Saini, Levi, Lorenz, Sauer, Boesel, et~al.]{esser2024scaling}
Patrick Esser, Sumith Kulal, Andreas Blattmann, Rahim Entezari, Jonas M{\"u}ller, Harry Saini, Yam Levi, Dominik Lorenz, Axel Sauer, Frederic Boesel, et~al.
\newblock Scaling rectified flow transformers for high-resolution image synthesis.
\newblock In \emph{Forty-first international conference on machine learning}, 2024.

\bibitem[Gao et~al.(2024)Gao, Zeng, Du, Cao, Zhou, Qi, Lai, So, Cao, Yang, et~al.]{gao2024seerattention}
Yizhao Gao, Zhichen Zeng, Dayou Du, Shijie Cao, Peiyuan Zhou, Jiaxing Qi, Junjie Lai, Hayden Kwok-Hay So, Ting Cao, Fan Yang, et~al.
\newblock Seerattention: Learning intrinsic sparse attention in your llms.
\newblock \emph{arXiv preprint arXiv:2410.13276}, 2024.

\bibitem[Gao et~al.(2025)Gao, Guo, Hoang, Huang, Jiang, Kong, Li, Li, Li, Li, et~al.]{gao2025seedance}
Yu~Gao, Haoyuan Guo, Tuyen Hoang, Weilin Huang, Lu~Jiang, Fangyuan Kong, Huixia Li, Jiashi Li, Liang Li, Xiaojie Li, et~al.
\newblock Seedance 1.0: Exploring the boundaries of video generation models.
\newblock \emph{arXiv preprint arXiv:2506.09113}, 2025.

\bibitem[Ghiasi et~al.(2018)Ghiasi, Lin, and Le]{ghiasi2018dropblock}
Golnaz Ghiasi, Tsung-Yi Lin, and Quoc~V Le.
\newblock Dropblock: A regularization method for convolutional networks.
\newblock \emph{Advances in neural information processing systems}, 31, 2018.

\bibitem[Huang et~al.(2025)Huang, Li, He, Zhou, and Shechtman]{huang2025self}
Xun Huang, Zhengqi Li, Guande He, Mingyuan Zhou, and Eli Shechtman.
\newblock Self forcing: Bridging the train-test gap in autoregressive video diffusion.
\newblock \emph{arXiv preprint arXiv:2506.08009}, 2025.

\bibitem[Jacobs et~al.(2023)Jacobs, Tanaka, Zhang, Zhang, Song, Rajbhandari, and He]{jacobs2023deepspeed}
Sam~Ade Jacobs, Masahiro Tanaka, Chengming Zhang, Minjia Zhang, Shuaiwen~Leon Song, Samyam Rajbhandari, and Yuxiong He.
\newblock Deepspeed ulysses: System optimizations for enabling training of extreme long sequence transformer models.
\newblock \emph{arXiv preprint arXiv:2309.14509}, 2023.

\bibitem[Jiang et~al.(2024)Jiang, Li, Zhang, Wu, Luo, Ahn, Han, Abdi, Li, Lin, et~al.]{jiang2024minference}
Huiqiang Jiang, Yucheng Li, Chengruidong Zhang, Qianhui Wu, Xufang Luo, Surin Ahn, Zhenhua Han, Amir Abdi, Dongsheng Li, Chin-Yew Lin, et~al.
\newblock Minference 1.0: Accelerating pre-filling for long-context llms via dynamic sparse attention.
\newblock \emph{Advances in Neural Information Processing Systems}, 37:\penalty0 52481--52515, 2024.

\bibitem[Kahatapitiya et~al.(2025)Kahatapitiya, Liu, He, Liu, Jia, Zhang, Ryoo, and Xie]{kahatapitiya2025adaptive}
Kumara Kahatapitiya, Haozhe Liu, Sen He, Ding Liu, Menglin Jia, Chenyang Zhang, Michael~S Ryoo, and Tian Xie.
\newblock Adaptive caching for faster video generation with diffusion transformers.
\newblock In \emph{Proceedings of the IEEE/CVF International Conference on Computer Vision}, pp.\  15240--15252, 2025.

\bibitem[Kong et~al.(2024)Kong, Tian, Zhang, Min, Dai, Zhou, Xiong, Li, Wu, Zhang, et~al.]{kong2024hunyuanvideo}
Weijie Kong, Qi~Tian, Zijian Zhang, Rox Min, Zuozhuo Dai, Jin Zhou, Jiangfeng Xiong, Xin Li, Bo~Wu, Jianwei Zhang, et~al.
\newblock Hunyuanvideo: A systematic framework for large video generative models.
\newblock \emph{arXiv preprint arXiv:2412.03603}, 2024.

\bibitem[Li et~al.(2024)Li, Lin, Zhang, Cai, Li, Guo, Xie, Meng, Zhu, and Han]{li2024svdquant}
Muyang Li, Yujun Lin, Zhekai Zhang, Tianle Cai, Xiuyu Li, Junxian Guo, Enze Xie, Chenlin Meng, Jun-Yan Zhu, and Song Han.
\newblock Svdquant: Absorbing outliers by low-rank components for 4-bit diffusion models.
\newblock \emph{arXiv preprint arXiv:2411.05007}, 2024.

\bibitem[Li et~al.(2025)Li, Li, Cai, Xi, Yang, Lin, Zhang, Yang, Hu, Peng, et~al.]{li2025radial}
Xingyang Li, Muyang Li, Tianle Cai, Haocheng Xi, Shuo Yang, Yujun Lin, Lvmin Zhang, Songlin Yang, Jinbo Hu, Kelly Peng, et~al.
\newblock Radial attention: O (nlog n) sparse attention with energy decay for long video generation.
\newblock \emph{arXiv preprint arXiv:2506.19852}, 2025.

\bibitem[Lin et~al.(2024)Lin, Ge, Cheng, Li, Zhu, Wang, He, Ye, Yuan, Chen, et~al.]{lin2024open}
Bin Lin, Yunyang Ge, Xinhua Cheng, Zongjian Li, Bin Zhu, Shaodong Wang, Xianyi He, Yang Ye, Shenghai Yuan, Liuhan Chen, et~al.
\newblock Open-sora plan: Open-source large video generation model.
\newblock \emph{arXiv preprint arXiv:2412.00131}, 2024.

\bibitem[Lipman et~al.()Lipman, Chen, Ben-Hamu, Nickel, and Le]{lipmanflow}
Yaron Lipman, Ricky~TQ Chen, Heli Ben-Hamu, Maximilian Nickel, and Matthew Le.
\newblock Flow matching for generative modeling.
\newblock In \emph{The Eleventh International Conference on Learning Representations}.

\bibitem[Liu et~al.(2023)Liu, Zaharia, and Abbeel]{liu2023ring}
Hao Liu, Matei Zaharia, and Pieter Abbeel.
\newblock Ring attention with blockwise transformers for near-infinite context.
\newblock \emph{arXiv preprint arXiv:2310.01889}, 2023.

\bibitem[Liu et~al.(2025)Liu, Wu, Zheng, Wei, He, Pi, and Chen]{liu2025videodpo}
Runtao Liu, Haoyu Wu, Ziqiang Zheng, Chen Wei, Yingqing He, Renjie Pi, and Qifeng Chen.
\newblock Videodpo: Omni-preference alignment for video diffusion generation.
\newblock In \emph{Proceedings of the Computer Vision and Pattern Recognition Conference}, pp.\  8009--8019, 2025.

\bibitem[Liu et~al.(2022)Liu, Gong, and Liu]{liu2022flow}
Xingchao Liu, Chengyue Gong, and Qiang Liu.
\newblock Flow straight and fast: Learning to generate and transfer data with rectified flow.
\newblock \emph{arXiv preprint arXiv:2209.03003}, 2022.

\bibitem[Lu et~al.(2025{\natexlab{a}})Lu, Jiang, Liu, Du, Jiang, Hong, Liu, He, Yuan, Wang, et~al.]{lu2025moba}
Enzhe Lu, Zhejun Jiang, Jingyuan Liu, Yulun Du, Tao Jiang, Chao Hong, Shaowei Liu, Weiran He, Enming Yuan, Yuzhi Wang, et~al.
\newblock Moba: Mixture of block attention for long-context llms.
\newblock \emph{arXiv preprint arXiv:2502.13189}, 2025{\natexlab{a}}.

\bibitem[Lu et~al.(2025{\natexlab{b}})Lu, Xia, Zhang, Kuang, Zheng, Ren, and Xiao]{lu2025hyper}
Yanzuo Lu, Xin Xia, Manlin Zhang, Huafeng Kuang, Jianbin Zheng, Yuxi Ren, and Xuefeng Xiao.
\newblock Hyper-bagel: A unified acceleration framework for multimodal understanding and generation.
\newblock \emph{arXiv preprint arXiv:2509.18824}, 2025{\natexlab{b}}.

\bibitem[Lv et~al.(2024)Lv, Si, Song, Yang, Qiao, Liu, and Wong]{lv2024fastercache}
Zhengyao Lv, Chenyang Si, Junhao Song, Zhenyu Yang, Yu~Qiao, Ziwei Liu, and Kwan-Yee~K Wong.
\newblock Fastercache: Training-free video diffusion model acceleration with high quality.
\newblock \emph{arXiv preprint arXiv:2410.19355}, 2024.

\bibitem[Ma et~al.(2024{\natexlab{a}})Ma, Fang, Bi~Mi, and Wang]{ma2024learning}
Xinyin Ma, Gongfan Fang, Michael Bi~Mi, and Xinchao Wang.
\newblock Learning-to-cache: Accelerating diffusion transformer via layer caching.
\newblock \emph{Advances in Neural Information Processing Systems}, 37:\penalty0 133282--133304, 2024{\natexlab{a}}.

\bibitem[Ma et~al.(2024{\natexlab{b}})Ma, Fang, and Wang]{ma2024deepcache}
Xinyin Ma, Gongfan Fang, and Xinchao Wang.
\newblock Deepcache: Accelerating diffusion models for free.
\newblock In \emph{Proceedings of the IEEE/CVF conference on computer vision and pattern recognition}, pp.\  15762--15772, 2024{\natexlab{b}}.

\bibitem[Mehta et~al.(2025)Mehta, Xin, Islam, Zhang, Baig, Goel, and Cavallari]{mehta2025tensorrt}
Gaurav Mehta, Jack Xin, Rifat Islam, Yashwant Zhang, Aadil Baig, Ankit Goel, and Stefano Cavallari.
\newblock Nvidia tensorrt unlocks fp4 image generation for nvidia blackwell geforce rtx 50 series gpus.
\newblock \url{https://developer.nvidia.com/blog/nvidia-tensorrt-unlocks-fp4-image-generation-for-nvidia-blackwell-geforce-rtx-50-series-gpus/}, May 2025.
\newblock NVIDIA Developer Blog.

\bibitem[Polyak et~al.(2025)Polyak, Zohar, Brown, Tjandra, Sinha, Lee, Vyas, Shi, Ma, Chuang, Yan, et~al.]{polyak2025moviegencastmedia}
Adam Polyak, Amit Zohar, Andrew Brown, Andros Tjandra, Animesh Sinha, Ann Lee, Apoorv Vyas, Bowen Shi, Chih-Yao Ma, Ching-Yao Chuang, David Yan, et~al.
\newblock Movie gen: A cast of media foundation models, 2025.
\newblock URL \url{https://arxiv.org/abs/2410.13720}.

\bibitem[Qiu et~al.(2025)Qiu, Wang, Zheng, Huang, Wen, Yang, Men, Yu, Huang, Huang, et~al.]{qiu2025gated}
Zihan Qiu, Zekun Wang, Bo~Zheng, Zeyu Huang, Kaiyue Wen, Songlin Yang, Rui Men, Le~Yu, Fei Huang, Suozhi Huang, et~al.
\newblock Gated attention for large language models: Non-linearity, sparsity, and attention-sink-free.
\newblock \emph{arXiv preprint arXiv:2505.06708}, 2025.

\bibitem[Salimans \& Ho(2022)Salimans and Ho]{salimans2022progressive}
Tim Salimans and Jonathan Ho.
\newblock Progressive distillation for fast sampling of diffusion models.
\newblock \emph{arXiv preprint arXiv:2202.00512}, 2022.

\bibitem[Shah et~al.(2024)Shah, Bikshandi, Zhang, Thakkar, Ramani, and Dao]{shah2024flashattention}
Jay Shah, Ganesh Bikshandi, Ying Zhang, Vijay Thakkar, Pradeep Ramani, and Tri Dao.
\newblock Flashattention-3: Fast and accurate attention with asynchrony and low-precision.
\newblock \emph{Advances in Neural Information Processing Systems}, 37:\penalty0 68658--68685, 2024.

\bibitem[Shazeer et~al.(2017)Shazeer, Mirhoseini, Maziarz, Davis, Le, Hinton, and Dean]{shazeer2017outrageously}
Noam Shazeer, Azalia Mirhoseini, Krzysztof Maziarz, Andy Davis, Quoc Le, Geoffrey Hinton, and Jeff Dean.
\newblock Outrageously large neural networks: The sparsely-gated mixture-of-experts layer.
\newblock \emph{arXiv preprint arXiv:1701.06538}, 2017.

\bibitem[Shen et~al.(2025)Shen, Li, Yang, Zhang, Zhang, Li, Wang, Lu, and Tang]{shen2025directly}
Xiangwei Shen, Zhimin Li, Zhantao Yang, Shiyi Zhang, Yingfang Zhang, Donghao Li, Chunyu Wang, Qinglin Lu, and Yansong Tang.
\newblock Directly aligning the full diffusion trajectory with fine-grained human preference.
\newblock \emph{arXiv preprint arXiv:2509.06942}, 2025.

\bibitem[Song \& Dhariwal(2023)Song and Dhariwal]{song2023improved}
Yang Song and Prafulla Dhariwal.
\newblock Improved techniques for training consistency models.
\newblock \emph{arXiv preprint arXiv:2310.14189}, 2023.

\bibitem[Song et~al.(2023)Song, Dhariwal, Chen, and Sutskever]{song2023consistency}
Yang Song, Prafulla Dhariwal, Mark Chen, and Ilya Sutskever.
\newblock Consistency models.
\newblock 2023.

\bibitem[Spector et~al.(2024)Spector, Arora, Singhal, Fu, and R{\'e}]{spector2024thunderkittens}
Benjamin~F Spector, Simran Arora, Aaryan Singhal, Daniel~Y Fu, and Christopher R{\'e}.
\newblock Thunderkittens: Simple, fast, and adorable ai kernels.
\newblock \emph{arXiv preprint arXiv:2410.20399}, 2024.

\bibitem[Su et~al.(2024)Su, Ahmed, Lu, Pan, Bo, and Liu]{su2024roformer}
Jianlin Su, Murtadha Ahmed, Yu~Lu, Shengfeng Pan, Wen Bo, and Yunfeng Liu.
\newblock Roformer: Enhanced transformer with rotary position embedding.
\newblock \emph{Neurocomputing}, 568:\penalty0 127063, 2024.

\bibitem[Tan et~al.(2025)Tan, Chen, Jiang, Chen, Yan, Duan, Zhu, Jiang, and Xu]{tan2025dsv}
Xin Tan, Yuetao Chen, Yimin Jiang, Xing Chen, Kun Yan, Nan Duan, Yibo Zhu, Daxin Jiang, and Hong Xu.
\newblock Dsv: Exploiting dynamic sparsity to accelerate large-scale video dit training, 2025.
\newblock URL \url{https://arxiv.org/abs/2502.07590}.

\bibitem[Tang et~al.(2024)Tang, Zhao, Zhu, Xiao, Kasikci, and Han]{tang2024quest}
Jiaming Tang, Yilong Zhao, Kan Zhu, Guangxuan Xiao, Baris Kasikci, and Song Han.
\newblock Quest: Query-aware sparsity for efficient long-context llm inference.
\newblock \emph{arXiv preprint arXiv:2406.10774}, 2024.

\bibitem[Team(2025)]{fastvideo2025posttraining}
FastVideo Team.
\newblock Fastwan: Generating a 5-second video in 5 seconds via sparse distillation.
\newblock \url{https://hao-ai-lab.github.io/blogs/fastvideo_post_training/}, August 2025.
\newblock Hao AI Lab @ UCSD Blog.

\bibitem[Wang et~al.(2025)Wang, Ai, Wen, Mao, Xie, Chen, Yu, Zhao, Yang, Zeng, et~al.]{wan2025wan}
Ang Wang, Baole Ai, Bin Wen, Chaojie Mao, Chen-Wei Xie, Di~Chen, Feiwu Yu, Haiming Zhao, Jianxiao Yang, Jianyuan Zeng, et~al.
\newblock Wan: Open and advanced large-scale video generative models.
\newblock \emph{arXiv preprint arXiv:2503.20314}, 2025.

\bibitem[Wang et~al.(2024)Wang, Huang, Bergman, Shen, Gao, Lingelbach, Sun, Bian, Song, Liu, et~al.]{wang2024phased}
Fu-Yun Wang, Zhaoyang Huang, Alexander Bergman, Dazhong Shen, Peng Gao, Michael Lingelbach, Keqiang Sun, Weikang Bian, Guanglu Song, Yu~Liu, et~al.
\newblock Phased consistency models.
\newblock \emph{Advances in neural information processing systems}, 37:\penalty0 83951--84009, 2024.

\bibitem[Wu et~al.(2025)Wu, Gao, Ye, Li, Li, Guo, Liu, Xue, Hou, Liu, et~al.]{wu2025rewarddance}
Jie Wu, Yu~Gao, Zilyu Ye, Ming Li, Liang Li, Hanzhong Guo, Jie Liu, Zeyue Xue, Xiaoxia Hou, Wei Liu, et~al.
\newblock Rewarddance: Reward scaling in visual generation.
\newblock \emph{arXiv preprint arXiv:2509.08826}, 2025.

\bibitem[Xi et~al.(2025)Xi, Yang, Zhao, Xu, Li, Li, Lin, Cai, Zhang, Li, Chen, Stoica, Keutzer, and Han]{xi2025sparsevideo}
Haocheng Xi, Shuo Yang, Yilong Zhao, Chenfeng Xu, Muyang Li, Xiuyu Li, Yujun Lin, Han Cai, Jintao Zhang, Dacheng Li, Jianfei Chen, Ion Stoica, Kurt Keutzer, and Song Han.
\newblock Sparse videogen: Accelerating video diffusion transformers with spatial-temporal sparsity, 2025.
\newblock URL \url{https://arxiv.org/abs/2502.01776}.

\bibitem[Xiao et~al.(2023)Xiao, Tian, Chen, Han, and Lewis]{xiao2023efficient}
Guangxuan Xiao, Yuandong Tian, Beidi Chen, Song Han, and Mike Lewis.
\newblock Efficient streaming language models with attention sinks.
\newblock \emph{arXiv preprint arXiv:2309.17453}, 2023.

\bibitem[Xu et~al.(2023)Xu, Liu, Wu, Tong, Li, Ding, Tang, and Dong]{xu2023imagereward}
Jiazheng Xu, Xiao Liu, Yuchen Wu, Yuxuan Tong, Qinkai Li, Ming Ding, Jie Tang, and Yuxiao Dong.
\newblock Imagereward: Learning and evaluating human preferences for text-to-image generation.
\newblock \emph{Advances in Neural Information Processing Systems}, 36:\penalty0 15903--15935, 2023.

\bibitem[Xu et~al.(2025)Xu, Xiao, Huang, Guo, and Han]{xu2025xattention}
Ruyi Xu, Guangxuan Xiao, Haofeng Huang, Junxian Guo, and Song Han.
\newblock Xattention: Block sparse attention with antidiagonal scoring.
\newblock \emph{arXiv preprint arXiv:2503.16428}, 2025.

\bibitem[Xue et~al.(2025)Xue, Wu, Gao, Kong, Zhu, Chen, Liu, Liu, Guo, Huang, et~al.]{xue2025dancegrpo}
Zeyue Xue, Jie Wu, Yu~Gao, Fangyuan Kong, Lingting Zhu, Mengzhao Chen, Zhiheng Liu, Wei Liu, Qiushan Guo, Weilin Huang, et~al.
\newblock Dancegrpo: Unleashing grpo on visual generation.
\newblock \emph{arXiv preprint arXiv:2505.07818}, 2025.

\bibitem[Yang et~al.()Yang, Sheng, Gonzalez, Stoica, and Zheng]{yang2408post}
S~Yang, Y~Sheng, JE~Gonzalez, I~Stoica, and L~Zheng.
\newblock Post-training sparse attention with double sparsity, 2024b.
\newblock \emph{URL https://arxiv.org/abs/2408.07092}.

\bibitem[Yang et~al.(2024)Yang, Teng, Zheng, Ding, Huang, Xu, Yang, Hong, Zhang, Feng, et~al.]{yang2024cogvideox}
Zhuoyi Yang, Jiayan Teng, Wendi Zheng, Ming Ding, Shiyu Huang, Jiazheng Xu, Yuanming Yang, Wenyi Hong, Xiaohan Zhang, Guanyu Feng, et~al.
\newblock Cogvideox: Text-to-video diffusion models with an expert transformer.
\newblock \emph{CoRR}, 2024.

\bibitem[Yin et~al.(2024{\natexlab{a}})Yin, Gharbi, Park, Zhang, Shechtman, Durand, and Freeman]{yin2024improved}
Tianwei Yin, Micha{\"e}l Gharbi, Taesung Park, Richard Zhang, Eli Shechtman, Fredo Durand, and William~T Freeman.
\newblock Improved distribution matching distillation for fast image synthesis.
\newblock In \emph{NeurIPS}, 2024{\natexlab{a}}.

\bibitem[Yin et~al.(2024{\natexlab{b}})Yin, Gharbi, Zhang, Shechtman, Durand, Freeman, and Park]{yin2024onestep}
Tianwei Yin, Micha{\"e}l Gharbi, Richard Zhang, Eli Shechtman, Fr{\'e}do Durand, William~T Freeman, and Taesung Park.
\newblock One-step diffusion with distribution matching distillation.
\newblock In \emph{CVPR}, 2024{\natexlab{b}}.

\bibitem[Yuan et~al.(2025)Yuan, Gao, Dai, Luo, Zhao, Zhang, Xie, Wei, Wang, Xiao, et~al.]{yuan2025native}
Jingyang Yuan, Huazuo Gao, Damai Dai, Junyu Luo, Liang Zhao, Zhengyan Zhang, Zhenda Xie, YX~Wei, Lean Wang, Zhiping Xiao, et~al.
\newblock Native sparse attention: Hardware-aligned and natively trainable sparse attention.
\newblock \emph{arXiv preprint arXiv:2502.11089}, 2025.

\bibitem[Yuan et~al.(2024)Yuan, Zhang, Pu, Ning, Zhang, Zhao, Yan, Dai, and Wang]{yuan2024ditfastattn}
Zhihang Yuan, Hanling Zhang, Lu~Pu, Xuefei Ning, Linfeng Zhang, Tianchen Zhao, Shengen Yan, Guohao Dai, and Yu~Wang.
\newblock Ditfastattn: Attention compression for diffusion transformer models.
\newblock \emph{Advances in Neural Information Processing Systems}, 37:\penalty0 1196--1219, 2024.

\bibitem[Zhan et~al.(2025)Zhan, Li, Shen, Zhang, Wu, and Zhang]{zhan2025bidirectional}
Chenlu Zhan, Wen Li, Chuyu Shen, Jun Zhang, Suhui Wu, and Hao Zhang.
\newblock Bidirectional sparse attention for faster video diffusion training.
\newblock \emph{arXiv preprint arXiv:2509.01085}, 2025.

\bibitem[Zhang et~al.(2024{\natexlab{a}})Zhang, Huang, Zhang, Wei, Zhu, and Chen]{zhang2024sageattention2}
Jintao Zhang, Haofeng Huang, Pengle Zhang, Jia Wei, Jun Zhu, and Jianfei Chen.
\newblock Sageattention2: Efficient attention with thorough outlier smoothing and per-thread int4 quantization.
\newblock \emph{arXiv preprint arXiv:2411.10958}, 2024{\natexlab{a}}.

\bibitem[Zhang et~al.(2024{\natexlab{b}})Zhang, Wei, Huang, Zhang, Zhu, and Chen]{zhang2024sageattention}
Jintao Zhang, Jia Wei, Haofeng Huang, Pengle Zhang, Jun Zhu, and Jianfei Chen.
\newblock Sageattention: Accurate 8-bit attention for plug-and-play inference acceleration.
\newblock \emph{arXiv preprint arXiv:2410.02367}, 2024{\natexlab{b}}.

\bibitem[Zhang et~al.(2025{\natexlab{a}})Zhang, Wang, Jiang, Yang, Zheng, Xi, Wang, Zhu, Zhao, Stoica, Gonzalez, Zhu, and Chen]{zhang2025sla}
Jintao Zhang, Haoxu Wang, Kai Jiang, Shuo Yang, Kaiwen Zheng, Haocheng Xi, Ziteng Wang, Hongzhou Zhu, Min Zhao, Ion Stoica, Joseph~E. Gonzalez, Jun Zhu, and Jianfei Chen.
\newblock Sla: Beyond sparsity in diffusion transformers via fine-tunable sparse-linear attention.
\newblock \emph{arXiv preprint arXiv:2509.24006}, 2025{\natexlab{a}}.

\bibitem[Zhang et~al.(2025{\natexlab{b}})Zhang, Wei, Zhang, Xu, Huang, Wang, Jiang, Zhu, and Chen]{zhang2025sageattention3}
Jintao Zhang, Jia Wei, Pengle Zhang, Xiaoming Xu, Haofeng Huang, Haoxu Wang, Kai Jiang, Jun Zhu, and Jianfei Chen.
\newblock Sageattention3: Microscaling fp4 attention for inference and an exploration of 8-bit training.
\newblock \emph{arXiv preprint arXiv:2505.11594}, 2025{\natexlab{b}}.

\bibitem[Zhang et~al.(2025{\natexlab{c}})Zhang, Xiang, Huang, Wei, Xi, Zhu, and Chen]{zhang2025sparge}
Jintao Zhang, Chendong Xiang, Haofeng Huang, Jia Wei, Haocheng Xi, Jun Zhu, and Jianfei Chen.
\newblock Spargeattn: Accurate sparse attention accelerating any model inference.
\newblock In \emph{International Conference on Machine Learning (ICML)}, 2025{\natexlab{c}}.

\bibitem[Zhang et~al.(2025{\natexlab{d}})Zhang, Chen, Huang, Lin, Liu, Stoica, Xing, and Zhang]{zhang2025vsa}
Peiyuan Zhang, Yongqi Chen, Haofeng Huang, Will Lin, Zhengzhong Liu, Ion Stoica, Eric Xing, and Hao Zhang.
\newblock Vsa: Faster video diffusion with trainable sparse attention.
\newblock \emph{arXiv preprint arXiv:2505.13389}, 2025{\natexlab{d}}.

\bibitem[Zhang et~al.(2025{\natexlab{e}})Zhang, Chen, Su, Ding, Stoica, Liu, and Zhang]{zhang2025fast}
Peiyuan Zhang, Yongqi Chen, Runlong Su, Hangliang Ding, Ion Stoica, Zhengzhong Liu, and Hao Zhang.
\newblock Fast video generation with sliding tile attention.
\newblock \emph{arXiv preprint arXiv:2502.04507}, 2025{\natexlab{e}}.

\bibitem[Zhang et~al.(2025{\natexlab{f}})Zhang, Chen, Su, Ding, Stoica, Liu, and Zhang]{zhang2025fastvideo}
Peiyuan Zhang, Yongqi Chen, Runlong Su, Hangliang Ding, Ion Stoica, Zhengzhong Liu, and Hao Zhang.
\newblock Fast video generation with sliding tile attention, 2025{\natexlab{f}}.
\newblock URL \url{https://arxiv.org/abs/2502.04507}.

\bibitem[Zhang et~al.(2025{\natexlab{g}})Zhang, Huang, Chen, Lin, Liu, Stoica, Xing, and Zhang]{zhang2025faster}
Peiyuan Zhang, Haofeng Huang, Yongqi Chen, Will Lin, Zhengzhong Liu, Ion Stoica, Eric~P Xing, and Hao Zhang.
\newblock Faster video diffusion with trainable sparse attention.
\newblock \emph{arXiv e-prints}, pp.\  arXiv--2505, 2025{\natexlab{g}}.

\bibitem[Zhang et~al.(2023)Zhang, Sheng, Zhou, Chen, Zheng, Cai, Song, Tian, R{\'e}, Barrett, et~al.]{zhang2023h2o}
Zhenyu Zhang, Ying Sheng, Tianyi Zhou, Tianlong Chen, Lianmin Zheng, Ruisi Cai, Zhao Song, Yuandong Tian, Christopher R{\'e}, Clark Barrett, et~al.
\newblock H2o: Heavy-hitter oracle for efficient generative inference of large language models.
\newblock \emph{Advances in Neural Information Processing Systems}, 36:\penalty0 34661--34710, 2023.

\bibitem[Zhao et~al.(2025)Zhao, Hong, Yang, Xiao, Li, Ling, Xie, Chen, Zhu, Zhang, et~al.]{zhao2025paroattention}
Tianchen Zhao, Ke~Hong, Xinhao Yang, Xuefeng Xiao, Huixia Li, Feng Ling, Ruiqi Xie, Siqi Chen, Hongyu Zhu, Yichong Zhang, et~al.
\newblock Paroattention: Pattern-aware reordering for efficient sparse and quantized attention in visual generation models.
\newblock \emph{arXiv preprint arXiv:2506.16054}, 2025.

\bibitem[Zhao et~al.(2023)Zhao, Gu, Varma, Luo, Huang, Xu, Wright, Shojanazeri, Ott, Shleifer, et~al.]{zhao2023pytorch}
Yanli Zhao, Andrew Gu, Rohan Varma, Liang Luo, Chien-Chin Huang, Min Xu, Less Wright, Hamid Shojanazeri, Myle Ott, Sam Shleifer, et~al.
\newblock Pytorch fsdp: experiences on scaling fully sharded data parallel.
\newblock \emph{arXiv preprint arXiv:2304.11277}, 2023.

\bibitem[Zheng et~al.(2024)Zheng, Peng, Yang, Shen, Li, Liu, Zhou, Li, and You]{zheng2024open}
Zangwei Zheng, Xiangyu Peng, Tianji Yang, Chenhui Shen, Shenggui Li, Hongxin Liu, Yukun Zhou, Tianyi Li, and Yang You.
\newblock Open-sora: Democratizing efficient video production for all.
\newblock \emph{arXiv preprint arXiv:2412.20404}, 2024.

\end{thebibliography}
\bibliographystyle{iclr2027_conference}

\clearpage
\appendix

\maketitlesupplementary

\section{More Qualitative Results}
\label{sec:more_qualitative_results}

\begin{figure}[H]
\centering
\includegraphics[width=\textwidth]{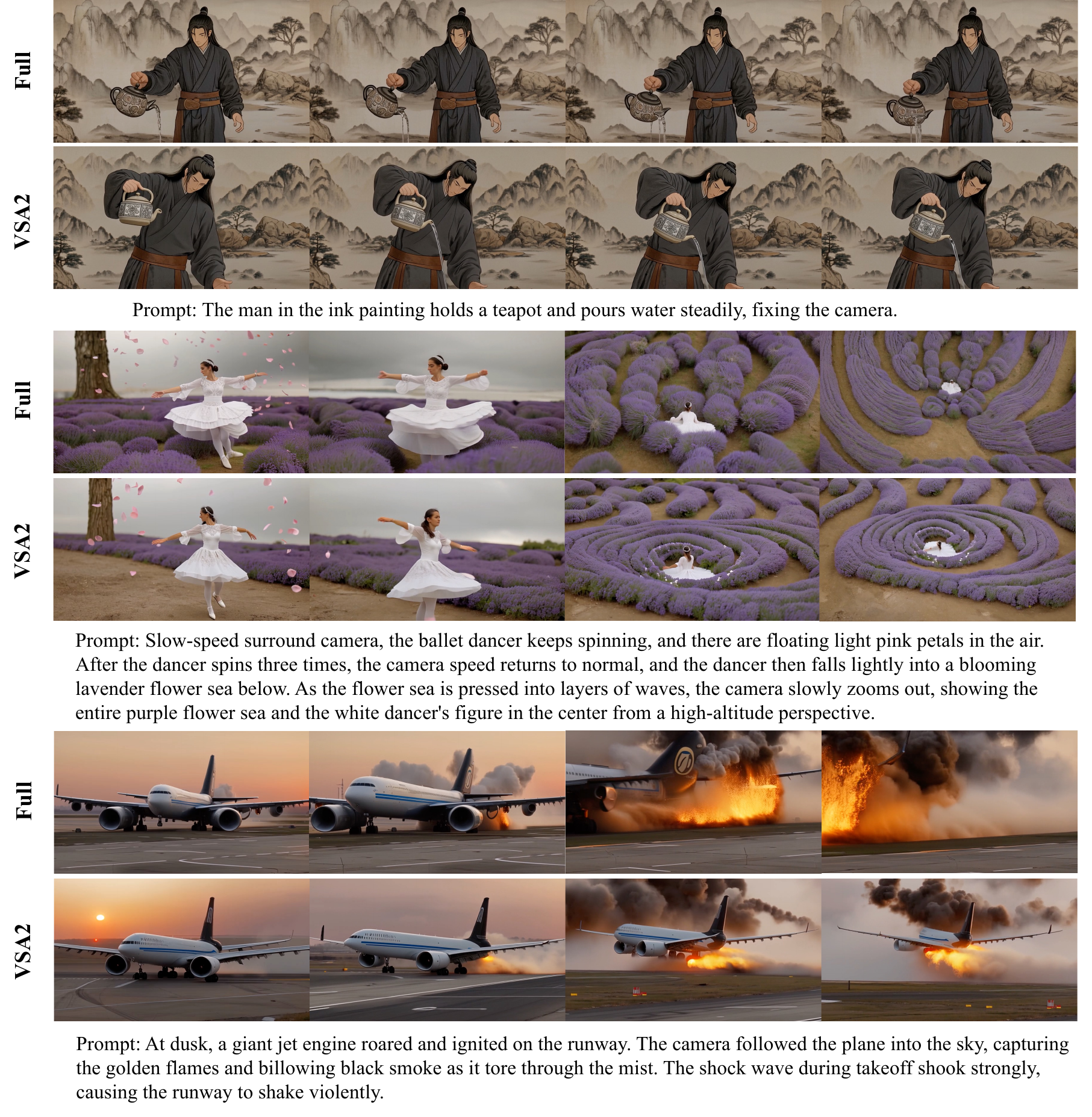}
\caption{More text-to-video samples of \methodnameshort~vs. full attention.}
\label{fig:qualitative_examples_t2v}
\end{figure}

\begin{figure}[H]
\centering
\includegraphics[width=\textwidth]{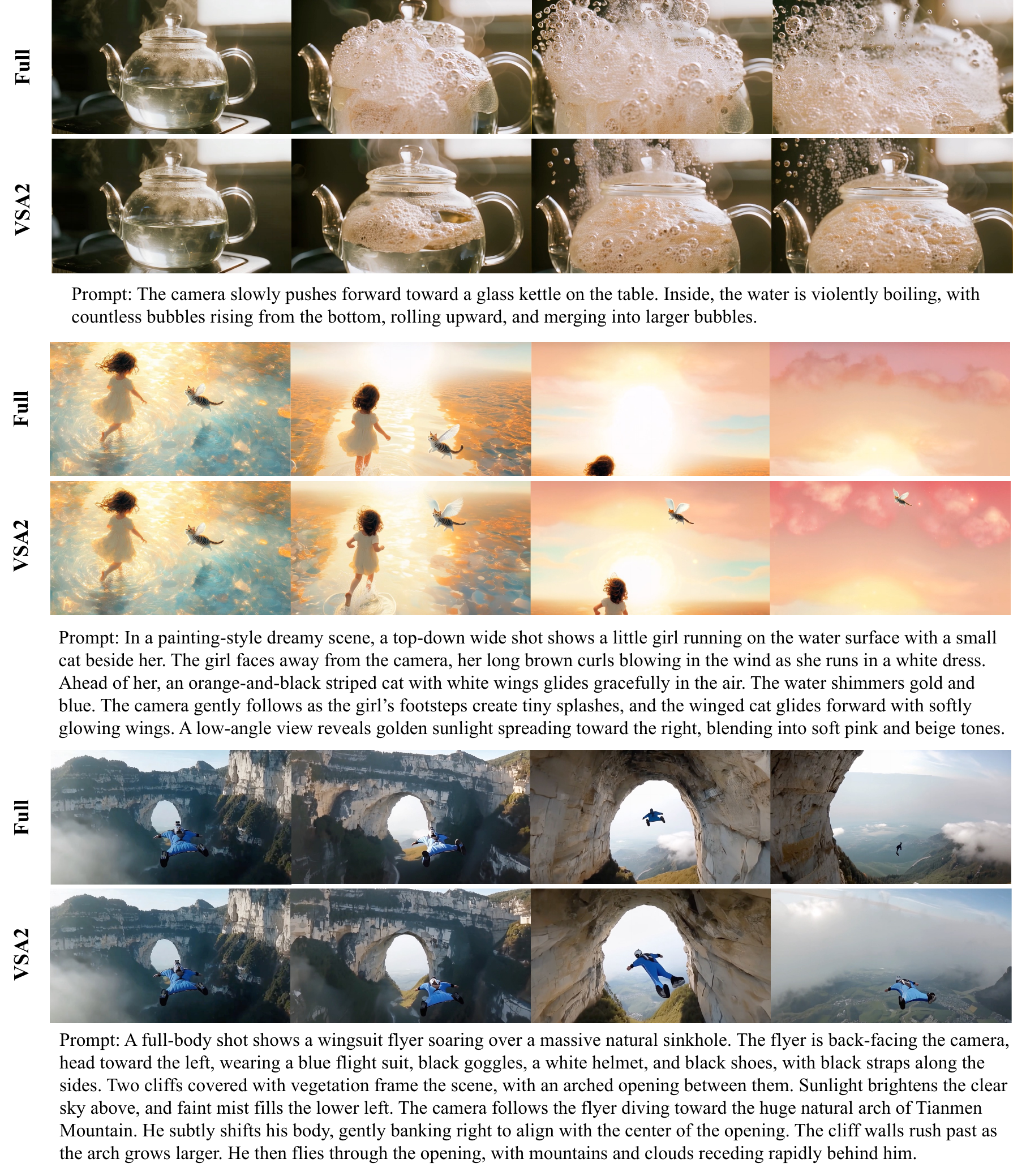}
\caption{More image-to-video samples of \methodnameshort~vs. full attention.}
\label{fig:qualitative_examples_i2v}
\end{figure}

\begin{figure}[H]
\centering
\includegraphics[width=\textwidth]{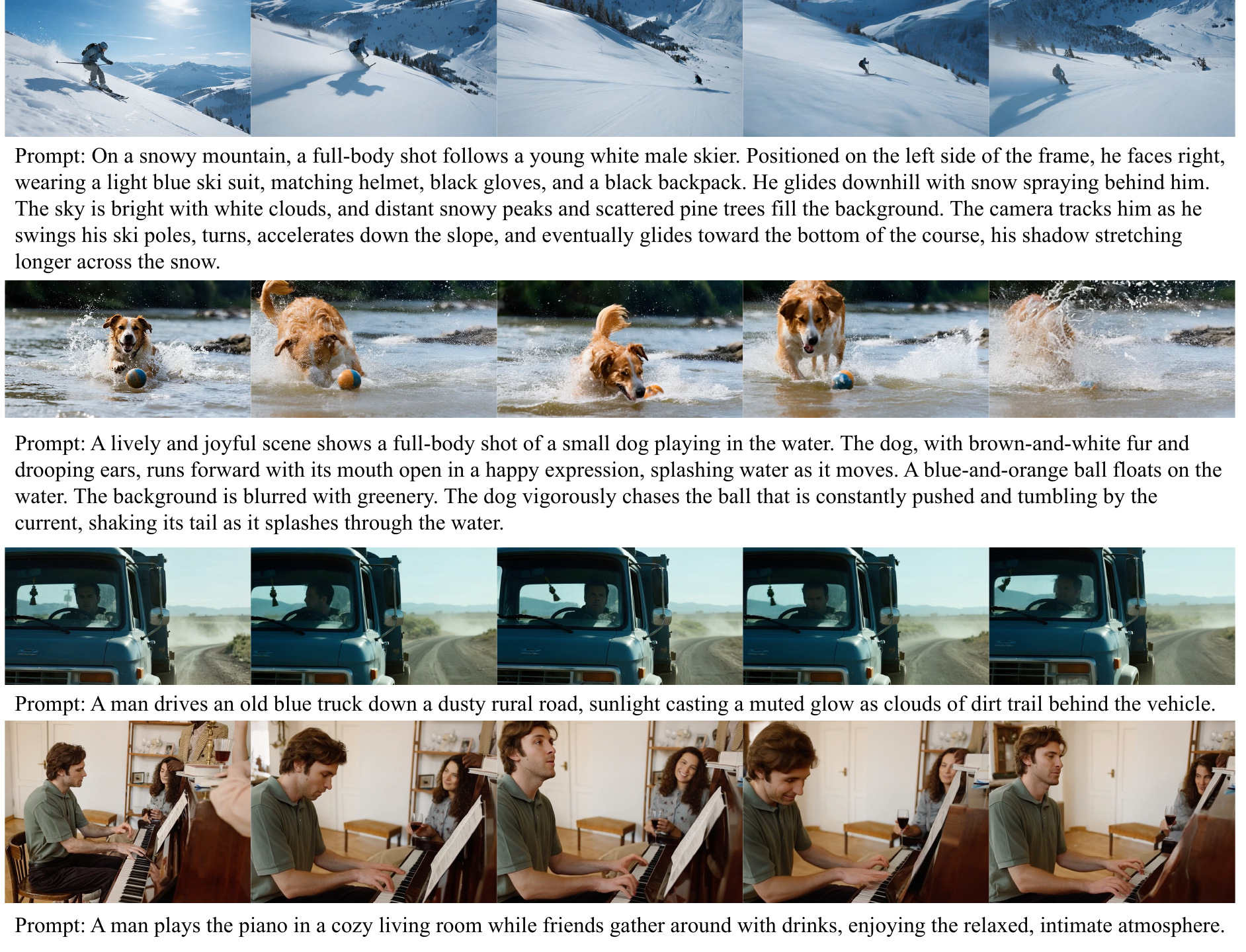}
\caption{Trained on 5--12\,s videos, \methodnameshort~can directly generate 30\,s videos.}
\label{fig:qualitative_examples_30sec}
\end{figure}

In Figures~\ref{fig:qualitative_examples_t2v}, \ref{fig:qualitative_examples_i2v}, and~\ref{fig:qualitative_examples_30sec}, we present more qualitative examples of \methodnameshort~vs. full attention generated from checkpoints in Exp.~6, Table~\ref{tab:main_t2v_i2v_human}. We also include video files in our supplementary material zip file.

\Needspace{6\baselineskip}
\section{Sparsity Computation}
\label{sec:sparsity_computation}

In this section, we describe how we compute the sparsity of
\methodnameshort~relative to full attention. Let $L$ denote the
sequence length, $B$ the block size, and $R$ the router input-pooling
size. The sequence contains $N=L/B$ blocks, assuming divisibility
for simplicity. We estimate the relative computational cost by
summing the dominant FLOPs of the constituent branches.

\begin{itemize}
    \item \textbf{Coarse Branch.}
    Pooling reduces the sequence length by a factor of $B$, so the
    coarse attention requires $1/B^2$ of the FLOPs of full attention.
    Under our configurations, this contribution and the overhead
    of gating, \texttt{tanh}, and pooling are negligible. We therefore
    omit them from the final sparsity estimate.

    \item \textbf{Router.}
    The router operates on pooled queries and keys of length $L/R$.
    Its dominant cost comes from two GEMM operations: one to compute
    the softmax normalization statistics and another to recompute
    the attention scores for softmax and score pooling.
    Together, these require approximately $1/R^2$ of the FLOPs of
    full attention, whose dominant cost likewise consists of two
    GEMMs.

    \item \textbf{Fine Branch.}
    The per-sequence topK selection retains $NK$ query--key block
    pairs in total, corresponding to $K$ selected KV blocks per
    query block on average. Each selected block pair contains
    $B^2$ token pairs. The fine branch therefore computes attention
    over $NKB^2$ token pairs, compared to $L^2$ for full attention,
    yielding a relative computational cost of
    \begin{equation}
        \frac{NKB^2}{L^2} = \frac{KB}{L}.
    \end{equation}
    Individual query blocks may attend to different numbers of
    KV blocks, while the total computation remains fixed by the
    sequence-level budget.
\end{itemize}

We define sparsity as one minus the computational cost relative
to full attention. Combining the router and fine-branch costs gives
\begin{equation}
    \text{Sparsity}
    \approx 1 - \left(\frac{1}{R^2} + \frac{KB}{L}\right).
\end{equation}
For fixed $B$, $R$, and $K$, the relative cost of the fine branch
decreases as the sequence length grows, while the relative router
cost remains constant.

\section{Reward Feedback Learning}
\label{sec:reward_feedback_learning}
We describe how we apply RL to pretrained checkpoints in Table~\ref{tab:main_t2v_i2v_human}. We use reward-feedback learning~\cite{xu2023imagereward} rather than DPO/PPO/GRPO variants. During training, the model predicts $x_0$ (the clean video), and multiple reward models evaluate the results. The objective maximizes a composite reward that combines a VLM-based RM with a CLIP-based RM. The gradients directly backpropagate through the reward model to video DiT. We tune the reward weights using the full-attention checkpoint and then reuse the same hyperparameters for \methodnameshort~without any adjustment, ensuring a fair comparison across models.

\section{Social Impacts}
\methodnameshort~reduces the training and inference cost of video diffusion models, making high-quality video generation more widely accessible. This can broaden access to creative tools across education, animation, and independent media production. At the same time, scalable realistic video generation introduces risks, including potential misuse for deepfakes or misleading content. We highlight the need to build strong detection systems and ethical safeguards alongside technical progress to ensure responsible deployment.

\Needspace{16\baselineskip}
\section{Future Work and Limitations}
We plan to apply \methodnameshort~to even larger sequence lengths, such as 1080p videos. Besides, \methodnameshort~is orthogonal to recent developments of autoregressive video generation methods such as Self-Forcing~\cite{huang2025self}, which we also plan to explore in the future. 
While the computation is fully balanced under the sequence parallelism strategy Ulysses~\cite{jacobs2023deepspeed}, the dynamic sparsity pattern of~\methodnameshort~may introduce compatibility challenges for Ring-Attention~\cite{liu2023ring}. Specifically, Ulysses performs sequence parallelism by gathering the full sequence on each GPU while sharding attention heads across devices. Since~\methodnameshort~maintains equal computation across attention heads, the workload remains naturally balanced. In contrast, Ring-Attention shards queries along the sequence dimension and overlaps attention computation with communication by sequentially gathering KV blocks. Because~\methodnameshort~does not enforce uniform KV selection across the sequence dimension, the selected KV blocks may become concentrated within a small subset of sequence partitions, leading to uneven computation workloads across GPUs. Another limitation is the router's fixed relative computational
cost, approximately $1/R^2$ of full attention, which could become
a bottleneck for extremely long sequences when sparsity is very high.

\end{document}